\documentclass[lettersize,journal]{IEEEtran}
\usepackage{amsmath,amsfonts}
\usepackage{algorithmic}
\usepackage{array}
\usepackage[caption=false,font=footnotesize,labelfont=sf,textfont=sf]{subfig}
\usepackage{xcolor}
\usepackage{textcomp}
\usepackage{stfloats}
\usepackage{url}
\usepackage{verbatim}
\usepackage{graphicx}
\usepackage{cite}
\usepackage{bm}
\usepackage{multirow}
\usepackage{ulem}
\usepackage{bbm}
\usepackage{booktabs}
\usepackage{makecell}
\usepackage[ruled,linesnumbered]{algorithm2e}
\usepackage[colorlinks,
            linkcolor=blue,
            anchorcolor=blue,
            citecolor=blue,  
            ]{hyperref}
\hypersetup{hidelinks,
	colorlinks=true,
	allcolors=black,
	pdfstartview=Fit,
        citecolor=black,
        anchorcolor=black,
	breaklinks=true}
\usepackage{cleveref}
\begin{document}

\sloppy
\title{When Heterophily Meets Heterogeneous Graphs: Latent Graphs Guided Unsupervised Representation Learning}

\author{Zhixiang Shen, and Zhao Kang
        % <-this % stops a space
% \thanks{Manuscript received April 19, 2021; revised August 16, 2021. This work was supported by the National Natural Science Foundation of China under Grant 62276053. \textit{(Corresponding author: Zhao Kang.)}}
\thanks{Z. Shen and Z. Kang are with the School of Computer Science and Engineering, University of Electronic Science and Technology of China, Chengdu 611731, China (e-mail: \href{mailto:zhixiang.zxs@gmail.com}{zhixiang.zxs@gmail.com}; \href{mailto:zkang@uestc.edu.cn}{zkang@uestc.edu.cn}).}}

\markboth{Journal of \LaTeX\ Class Files,~Vol.~14, No.~8, August~2021}%
{Shell \MakeLowercase{\textit{et al.}}: A Sample Article Using IEEEtran.cls for IEEE Journals}

\IEEEpubid{0000--0000/00\$00.00~\copyright~2021 IEEE}

\maketitle

\begin{abstract}
Unsupervised heterogeneous graph representation learning (UHGRL) has gained increasing attention due to its significance in handling practical graphs without labels. However, heterophily has been largely ignored, despite its ubiquitous presence in real-world heterogeneous graphs. In this paper, we define semantic heterophily and propose an innovative framework called Latent Graphs Guided Unsupervised Representation Learning (LatGRL) to handle this problem. First, we develop a similarity mining method that couples global structures and attributes, enabling the construction of fine-grained homophilic and heterophilic latent graphs to guide the representation learning. Moreover, we propose an adaptive dual-frequency semantic fusion mechanism to address the problem of node-level semantic heterophily. To cope with the massive scale of real-world data, we further design a scalable implementation. Extensive experiments on benchmark datasets validate the effectiveness and efficiency of our proposed framework. The source code and datasets have been made available at \href{https://github.com/zxlearningdeep/LatGRL}{https://github.com/zxlearningdeep/LatGRL}.
\end{abstract}


\section{Introduction}\label{1}

\IEEEPARstart{H}{eterogeneous} graphs, prevalent in various domains such as social networks, bibliographic networks, and knowledge graphs, represent complex semantic relationships \cite{li2021joint}. In a heterogeneous graph, nodes represent entities of multiple types and edges demonstrate various kinds of relations. 
Most cutting-edge methods model heterogeneous graphs based on meta-paths \cite{bing2023heterogeneous}. These techniques employ predefined meta-paths to extract a series of homogeneous graphs that pertain to a specific type of nodes for subsequent node representation learning, which aligns with the principles of multi-view graph learning \cite{pan2021multi}.
Recently, Unsupervised Heterogeneous Graph Representation Learning (UHGRL) has gained considerable attention due to its importance in handling large amounts of real-world graphs \cite{xie2021survey}. UHGRL leverages unsupervised techniques, such as contrastive learning, to obtain high-quality node representations, avoiding the reliance on labeled data. These representations aid in tasks such as node classification and clustering, finding practical applications in recommendation systems, and bibliographic network mining \cite{liu2022graph}.

\begin{figure}[t]
	\centering
		\includegraphics[width=0.9\linewidth]{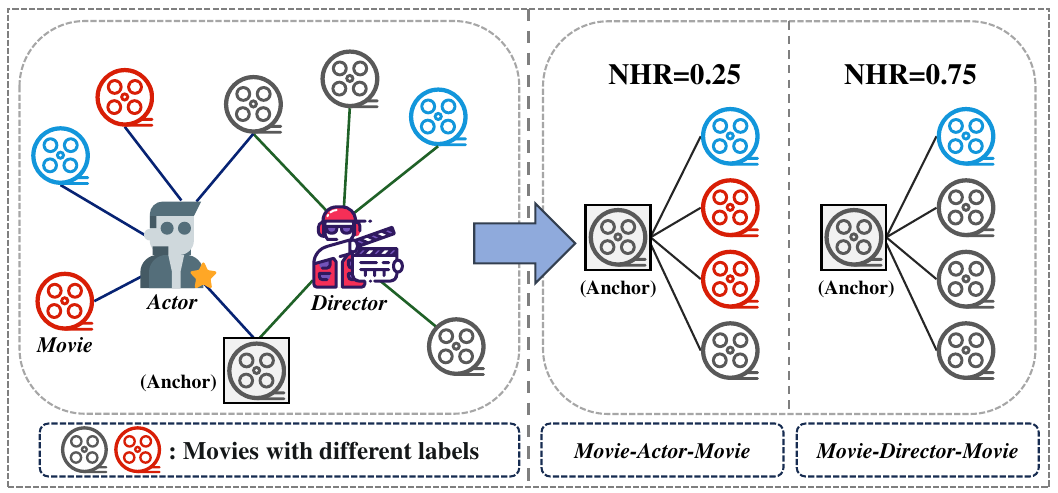}
	\caption{An example of semantic heterophily. The anchor node has distinct node-level semantic homophily ratios (NHR) across different meta-paths.}
	\label{SEMANTIC_HETEROPHILY}
\end{figure}

However, the semantic heterophily, which refers to the phenomenon that nodes of the same type connected by meta-paths often present dissimilar attributes or carry different labels, remains largely overlooked in current research despite its prominent existence in practical graphs. 
As shown in Fig. \ref{SEMANTIC_HETEROPHILY}, within the movie website, it is evident that the same actor may participate in films of various genres while a single director may also helm different genres of movies, reflecting semantic heterophily. Moreover, the anchor node demonstrates diverse node-level homophily ratios across distinct meta-paths.
In heterogeneous graphs, the presence of multiple relations amplifies the complexity of semantic heterophily. However, existing UHGRL methods, which are largely based on encoding mechanisms incorporating low-pass filtering and contrastive views with pronounced structural dependency \cite{park2020unsupervised,wang2021self,yu2023heterogeneous,tian2023heterogeneous}, tend to blur the representations of nearby nodes belonging to different categories. This limitation hinders their adaptability to semantic heterophily scenarios and compromises the discriminative capability of the representations. Consequently, two key problems need to be addressed: $(\mathcal{Q}1)$ \textit{How to quantitatively characterize semantic heterophily in heterogeneous graphs?} $(\mathcal{Q}2)$ \textit{ How to design an effective UHGRL framework to tackle semantic heterophily?}

\IEEEpubidadjcol

In this paper, we conduct an extensive study on $(\mathcal{Q}1)$. We propose two evaluation metrics: meta-path-level semantic homophily ratio (MHR) and node-level semantic homophily ratio (NHR). Through empirical analysis, we discover that real-world heterogeneous graphs exhibit diverse neighborhood patterns. Specifically, within the same meta-path, different nodes display a variety of NHR. Additionally, certain nodes may have low NHR under one meta-path, while higher NHR under another meta-path. The complex neighborhood patterns pose significant challenges for node representation learning.

To address $(\mathcal{Q}2)$, we propose a new UHGRL framework named Latent Graphs Guided Unsupervised Representation Learning (LatGRL). Divergent from existing graph contrastive methods that rely solely on contrastive views of the original topological structure \cite{you2020graph,wang2021self}, we present a new similarity mining approach that couples global structures and node attributes to construct homophilic and heterophilic latent graphs. Furthermore, we introduce an adaptive dual-frequency semantic fusion mechanism incorporating dual-pass graph filtering, which can simultaneously handle complex homophilic and heterophilic neighborhood patterns and facilitate node-wise modeling.
Finally, we exploit the meticulously constructed latent graphs as category-guided information to guide the representation learning process. 
In the context of the homophilic latent graph, the neighboring nodes predominantly belong to the same category, thereby facilitating the acquisition of shared information about that specific category.
On the contrary, the heterophilic latent graph exhibits a notable presence of neighbors from diverse categories, which provides valuable guidance by exposing the distinct characteristics between different categories. Our approach of concurrent guidance from homophilic and heterophilic latent graphs embodies significant innovation in the existing domains of heterogeneous and homogeneous graph learning. To ensure scalability and computational efficiency in large-scale graphs, we further optimize LatGRL to improve practical applicability.

Our contributions can be summarized as follows.

\begin{itemize}
\item To our knowledge, this is the first work considering semantic heterophily in unsupervised heterogeneous graph learning. We define semantic heterophily and explore complex node neighborhood patterns through in-depth empirical study.

\item We propose a novel framework to address the challenges of semantic heterophily. LatGRL employs a similarity mining approach that couples global structures and features and constructs homophilic and heterophilic latent graphs to guide representation learning. An adaptive dual-frequency semantic fusion mechanism with dual-pass graph filtering is proposed to handle the various patterns of the node neighborhoods.

\item A scalable method is further developed for large-scale data. Extensive experiments in classification and clustering demonstrate the effectiveness and efficiency of our approach.
\end{itemize}

\section{Preliminary}\label{2}

\subsection{Unsupervised Heterogeneous Graph Representation Learning}

A heterogeneous graph can be defined as a graph $\mathcal{G} = (\mathcal{V}, \mathcal{E}, \phi, \psi)$, where $\mathcal{V}$ is the node set and $\mathcal{E}$ is the edge set. $\phi : \mathcal{V}\rightarrow\mathcal{T}$ is the node-type mapping function where $\mathcal{T} = \{\phi(v) : v \in \mathcal{V}\}$, and $\psi : \mathcal{E}\rightarrow\mathcal{R}$ is the edge-type mapping function where $\mathcal{R} = \{\psi(e) : e \in \mathcal{E}\}$.

\noindent \textbf{Definition 1. Mete-path.} A meta-path $\mathcal{P} : \mathcal{T}_1\xrightarrow{\mathcal{R}_1}\mathcal{T}_2\xrightarrow{\mathcal{R}_2}\cdots\xrightarrow{\mathcal{R}_l}\mathcal{T}_{l+1}$ (abbreviated as $\mathcal{T}_1\mathcal{T}_2\cdots \mathcal{T}_{l+1}$) defines a composition relation $\mathcal{R}=\mathcal{R}_1\circ\mathcal{R}_2\circ\cdots\circ\mathcal{R}_l$ between nodes of type $\mathcal{T}_1$ and $\mathcal{T}_{l+1}$, where $\circ$ denotes the composition operator.

\noindent \textbf{Definition 2. Mete-path Subgraph.} A meta-path subgraph is defined as a graph $\mathcal{G}_{\Phi} = (\mathcal{V}_{\Phi}, \mathcal{E}_{\Phi})$ induced by the meta-path $\Phi = \mathcal{T}_1\mathcal{T}_2\cdots\mathcal{T}_{l+1}$. The meta-path subgraph becomes a homogeneous graph with edges in the relation defined by the meta-path $\Phi$ if $\mathcal{T}_1=\mathcal{T}_{l+1}$, which is usually used in meta-path-based HG algorithms.

\begin{table}[t]
\centering
\begin{minipage}{\linewidth} 
\caption{Frequently used notations.}
\centering
\resizebox{1.0\linewidth}{!}
{
\begin{tabular}{l|l}
\toprule
\textbf{Notation }                                   & \textbf{Description}                                     \\
\midrule
$\mathcal{G}  $                         & The heterogeneous graph.                      \\
$\mathcal{G}_{\Phi}$                                         & The homogeneous mete-path subgraph.                            \\
$P, N, d_f$                                         & The number of meta-paths/target nodes/attributes. \\
$\mathbf{X}$          & The attribute matrix of target nodes.                      \\
$\mathbf{X}_{i\cdot}$         & The attribute vector of node $v_i$.                      \\
$\mathbf{A}^{\Phi},\mathbf{L}^{\Phi}$              & The Adjacency/Laplacian matrix of $\mathcal{G}_{\Phi}$. \\  
$\mathcal{N}_i^{\Phi}$ & The neighbors set of node $v_i$ based on meta-path $\Phi$. \\
$y_i$          & The label of node $v_i$.                      \\
\midrule
$\mathbf{M}^{\Phi}$ & The random walk normalized adjacency matrix of $\mathcal{G}_{\Phi}$. \\
$sim(v_i, v_j)$ & The coupled similarity between two nodes.\\
$\mathbf{A}^S,\mathbf{A}^W$ & The adjacency matrices of homophilic/heterophilic latent graphs. \\
$\mathbf{H}^{\Phi, l}, \mathbf{H}^{\Phi, h} $ & The low-frequency/high-frequency node representations of $\mathcal{G}_{\Phi}$. \\
$\mathbf{Z}$ & The semantic-fusion node representations. \\
$\mathbf{Z}^l,\mathbf{Z}^h$ & The node representations of homophilic/heterophilic latent graphs. \\
$r$ & The filter order. \\
$K$ & The number of neighbors in latent graphs. \\
$k_{pos}$ & The number of positive samplers. \\
$\gamma$ & The sharpening hyper-parameter in $\mathcal{L}_{SCE}$. \\
$\tau$ & The temperature hyper-parameter in $\mathcal{L}_{C}$. \\
\midrule
$H(\mathcal{G}_{\Phi})$ & The Meta-path-level Semantic Homophily ratio (MHR). \\
$h(v_i)_{\Phi}$ & The Node-level Semantic Homophily ratio (NHR). \\
$\mathbbm{P}_i$ & The positive samples set of node $v_i$. \\
$I(\mathbf{Z},\mathbf{Z}^l)$ & The mutual information between $\mathbf{Z}$ and $\mathbf{Z}^l$. \\
$\mathcal{L}$ & The overall loss function. \\
\midrule

$\odot $ & The Hadamard product. \\
$\sigma(\cdot)$ & The non-linear activation function. \\
$[\cdot\parallel\cdot]$ & The concatenation operation. \\
\bottomrule
\end{tabular}
}
\label{Notations}
\end{minipage}
\end{table}

UHGRL aims to learn low-dimensional node representations without the supervision of labels.
In this paper, we adopt the previous task setting that focuses solely on a specific type of nodes \cite{wang2021self}, denoted as target nodes, which are used in downstream tasks such as node classification and clustering. Our proposed framework is rooted in the foundation of meta-path, expanding its horizons to encompass the vast domain of multi-view graph learning.

Given a heterogeneous graph $\mathcal{G}$ with node attribute matrix $\mathbf{X}\in\mathbb{R}^{N\times d_f}$, where $N$ is the number of target nodes, we define $\mathbf{X}_{i\cdot}\in\mathbb{R}^{d_f}$ as the feature vector of $i^{th}$ node and $\bm{x}\in\mathbb{R}^N$ is a column of the feature matrix representing a graph signal. Based on a set of meta-paths, we denote $\mathbf{A}^{\Phi}$ as the symmetric adjacency matrix of homogeneous subgraph $\mathcal{G}_{\Phi}$ induced by the meta-path $\Phi$. $\mathbf{D}_{\Phi}$ is the degree matrix of $\mathcal{G}_{\Phi}$ with ${\mathbf{D}_{\Phi}}_{ii} = \textstyle\sum_j\mathbf{A}^{\Phi}_{ij}$ and $\mathbf{L}^{\Phi}$ is the corresponding Laplacian matrix defined as $\mathbf{L}^{\Phi} = \mathbf{D}_{\Phi} - \mathbf{A}^{\Phi}$. 
Furthermore, the renormalized version of the adjacency matrix with self-loop is defined as $\tilde{\mathbf{A}}^{\Phi}_{sym} = \tilde{\mathbf{D}}_{\Phi}^{-\frac{1}{2}}\tilde{\mathbf{A}}^{\Phi}\tilde{\mathbf{D}}_{\Phi}^{-\frac{1}{2}}$ and the corresponding renormalized Laplacian matrix is denoted by $\tilde{\mathbf{L}}^{\Phi}_{sym} = \tilde{\mathbf{D}}_{\Phi}^{-\frac{1}{2}}\tilde{\mathbf{L}}^{\Phi}\tilde{\mathbf{D}}_{\Phi}^{-\frac{1}{2}} = \mathbf{I} - \tilde{\mathbf{A}}^{\Phi}_{sym}$, where $\tilde{\mathbf{A}}^{\Phi} = \mathbf{A}^{\Phi} + \mathbf{I}$, $\tilde{\mathbf{L}}^{\Phi} = \mathbf{L}^{\Phi} + \mathbf{I}$, and $\tilde{\mathbf{D}}_{\Phi} = \mathbf{D}_{\Phi} + \mathbf{I}$. More frequently used notations are summarized in Table \ref{Notations}.

\subsection{Homophily versus Heterophily}\label{2-B}
Two nodes are considered similar if they share the same label \cite{homo-necessity}. In a homogeneous graph, edges connecting two nodes with the same label are homophilic, while edges connecting nodes of different labels are heterophilic. Therefore, homogeneous graphs can be categorized into homophilic (homophily) graphs and heterophilic (heterophily) graphs on the basis of the proportion of homophilic edges. The edge level homophily ratio (HR) is usually defined as \cite{zhu2020beyond}:

\begin{equation}
H(\mathcal{G}) = \frac{1}{\mid\mathcal{E}\mid}\sum\limits_{(v_i,v_j)\in\mathcal{E}}\mathbbm{1}(y_i=y_j)
\end{equation}
where $\mid\mathcal{E}\mid$ is the size of edge set, $y_i$ is the label of node $v_i$ and $\mathbbm{1}(\cdot)$ denotes the indicator function (i.e., $\mathbbm{1}(\cdot)=1$ if the condition holds, otherwise $\mathbbm{1}(\cdot)=0$). A graph is considered homophilic when the edge level homophily ratio is large (typically, $0.5 \le H(\cdot) \le 1$); otherwise, it is a heterophilic graph.

\subsection{Graph Filtering}\label{2-C}

From a spectral perspective, the Laplacian filter amplifies the high-frequency components and suppresses the low-frequency components in the graph, while affinity matrices, such as the renormalized adjacency matrix, exhibit the opposite behavior \cite{hoang2021revisiting}. On the other hand, from a spatial perspective, the application of filters $\tilde{\mathbf{A}}_{sym}$ and $\tilde{\mathbf{L}}_{sym}$ to the graph signal $\bm{x} \in \mathbbm{R}^{N}$ can be interpreted as operations of aggregation and diversification. Formally:

\begin{equation}
(\tilde{\mathbf{A}}_{sym}\bm{x})_i = \frac{1}{\mid\mathcal{N}_i\mid}\sum\limits_{j\in\mathcal{N}_i}\bm{x}_j
\end{equation}
\begin{equation}
(\tilde{\mathbf{L}}_{sym}\bm{x})_i = \frac{1}{\mid\mathcal{N}_i\mid}\sum\limits_{j\in\mathcal{N}_i}(\bm{x}_i - \bm{x}_j)
\end{equation}
where $\mathcal{N}_i$ is the neighbor set of node $v_i$ with self-loop. Therefore, these two operations aim to smooth and sharpen the node features, respectively, and then capture the commonalities and differences among neighborhoods \cite{luan2022complete}. Most existing GNN encoders are essentially equivalent to low-pass graph filters \cite{bo2021beyond}, which limits their effectiveness to graphs with high homophily levels. Some research has attempted to introduce high-pass filters to handle heterophilic graphs \cite{luan2022revisiting}. These filters can sharpen the features of nodes across heterophilic edges, thereby avoiding the ambiguity of representations between nodes of different categories.

\begin{table}[]
\centering
\caption{The HR ($\%$) of the KNN graphs compared to the MHR ($\%$) of the original heterogeneous graphs.}
\resizebox{\linewidth}{!}
{
\renewcommand{\arraystretch}{1.25}
\begin{tabular}{|c|c|c|cc|}
\hline
\multirow{2}{*}{Dataset} & \multirow{2}{*}{Node Type}                                                                       & \multirow{2}{*}{MHR}                                                             & \multicolumn{2}{c|}{\begin{tabular}[c]{@{}c@{}}HR of \\ KNN Graph\end{tabular}} \\ \cline{4-5} 
                         &                                                                                                  &                                                                                  & \multicolumn{1}{c|}{K=5}                          & K=10                        \\ \hline
ACM                      & \begin{tabular}[c]{@{}c@{}}Paper (P), \\ Author (A), Subject (S)\end{tabular}                    & \begin{tabular}[c]{@{}c@{}}PAP: 80.85\\ PSP: 63.93\end{tabular}                  & \multicolumn{1}{c|}{74.17}                        & 72.01                       \\ \hline
IMDB                     & \begin{tabular}[c]{@{}c@{}}Movie (M), \\ Director (D), Author (A)\end{tabular}                   & \begin{tabular}[c]{@{}c@{}}MDM: 61.41\\ MAM: 44.43\end{tabular}                  & \multicolumn{1}{c|}{46.91}                        & 45.43                       \\ \hline
DBLP                     & \begin{tabular}[c]{@{}c@{}}Author (A), Paper (P), \\ Conference (C), Term (T)\end{tabular}       & \begin{tabular}[c]{@{}c@{}}APA: 79.88\\ APCPA: 66.97\\ APTPA: 32.45\end{tabular} & \multicolumn{1}{c|}{68.39}                        & 65.91                       \\ \hline
Yelp                     & \begin{tabular}[c]{@{}c@{}}Business (B), Service (S),\\ User (U), Rating Levels (L)\end{tabular} & \begin{tabular}[c]{@{}c@{}}BSB: 64.08\\ BUB: 44.97\\ BLB: 38.76\end{tabular}     & \multicolumn{1}{c|}{88.69}                        & 88.25                       \\ \hline
\end{tabular}
}
\label{HR-TABLE}%
\end{table}

\section{Empirical Study}\label{4}

We propose the notion of \textit{\textbf{Semantic Homophily}} in heterogeneous graphs, which refers to \textit{the tendency for nodes of the same type, connected by meta-paths, to possess similar features or identical labels. Conversely, when nodes of the same type exhibit dissimilar features or different labels, we refer to it as \textbf{Semantic Heterophily}}. To manifest the distinctions among different semantic relations, we calculate the homophily ratio separately for each homogeneous meta-path subgraph extracted from different meta-paths and refer to it as the meta-path-level semantic homophily ratio.

\noindent \textbf{Definition 3. Meta-path-level Semantic Homophily ratio (MHR).} Given a meta-path $\Phi = \mathcal{T}_1\mathcal{T}_2\cdots\mathcal{T}_{l+1}$ with $\mathcal{T}_1 = \mathcal{T}_{l+1}$ and the node label vector $y$, we define the meta-path-level semantic homophily ratio as:

\begin{equation}
H(\mathcal{G}_{\Phi}) = \frac{1}{\mid\mathcal{E}_{\Phi}\mid}\sum\limits_{(v_i,v_j)\in\mathcal{E}_{\Phi}}\mathbbm{1}(y_i=y_j)
\end{equation}
where $\mid\mathcal{E}_{\Phi}\mid$ is the number of edges in $\mathcal{G}_{\Phi}$. With the aid of MHR, we comprehensively evaluate the semantic homophily levels of four popular heterogeneous graph datasets in Table \ref{HR-TABLE}. Note that many graph contrastive learning methods rely on the K Nearest Neighbors (KNN) algorithm applied to node features for positive sampling or the contrastive view constructing 
 \cite{chen2023attribute,pan2021multi}. Therefore, we also assess the HR of the KNN graph, where cosine distance is adopted to measure distance.

The results demonstrate that different relations from the same heterogeneous graph exhibit varying degrees of semantic homophily. Specifically, certain meta-paths tend to connect nodes of the same class, while others are more likely to connect nodes of different classes. Moreover, it is intriguing that the positive sampling technique, KNN, is not universally reliable. Except for Yelp, the HR of the KNN graph is consistently lower than that of the highest meta-path subgraph in the original graph. Furthermore, as K increases, the HR of the KNN graph decreases further. However, MHR overlooks the disparities between different nodes. So, we define the semantic homophily ratio at the node level.

\begin{figure*}[!t]
\centering
\subfloat[ACM (1D)\label{ACM-1D}]{\includegraphics[width=0.3\linewidth]{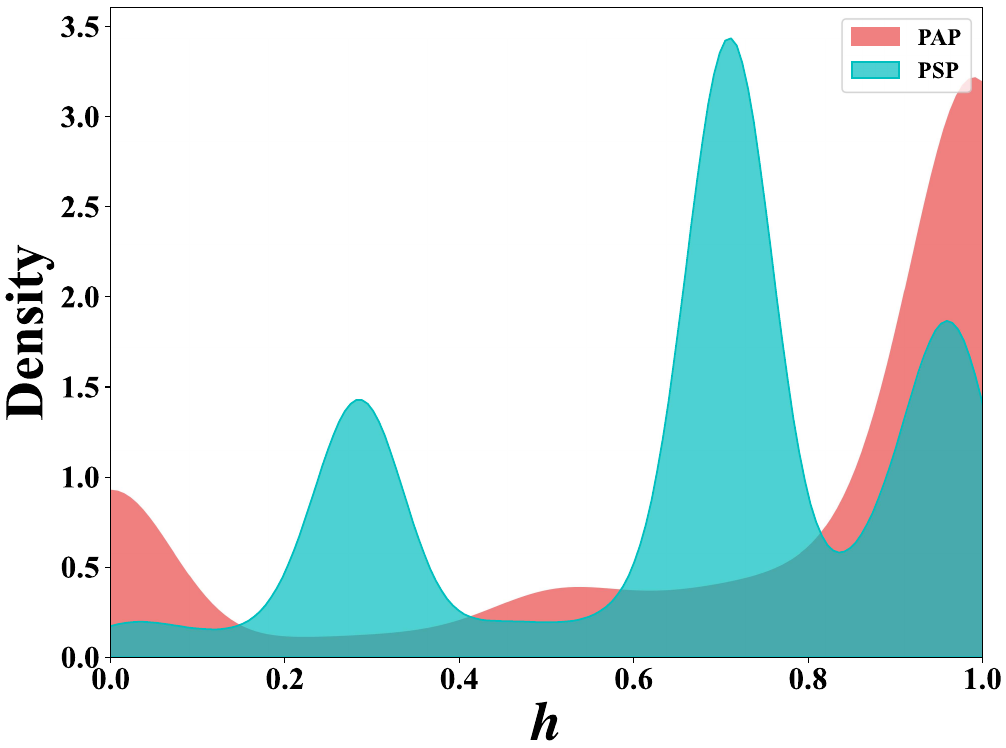}}
\hfil
\subfloat[DBLP (1D)\label{DBLP-1D}]{\includegraphics[width=0.3\linewidth]{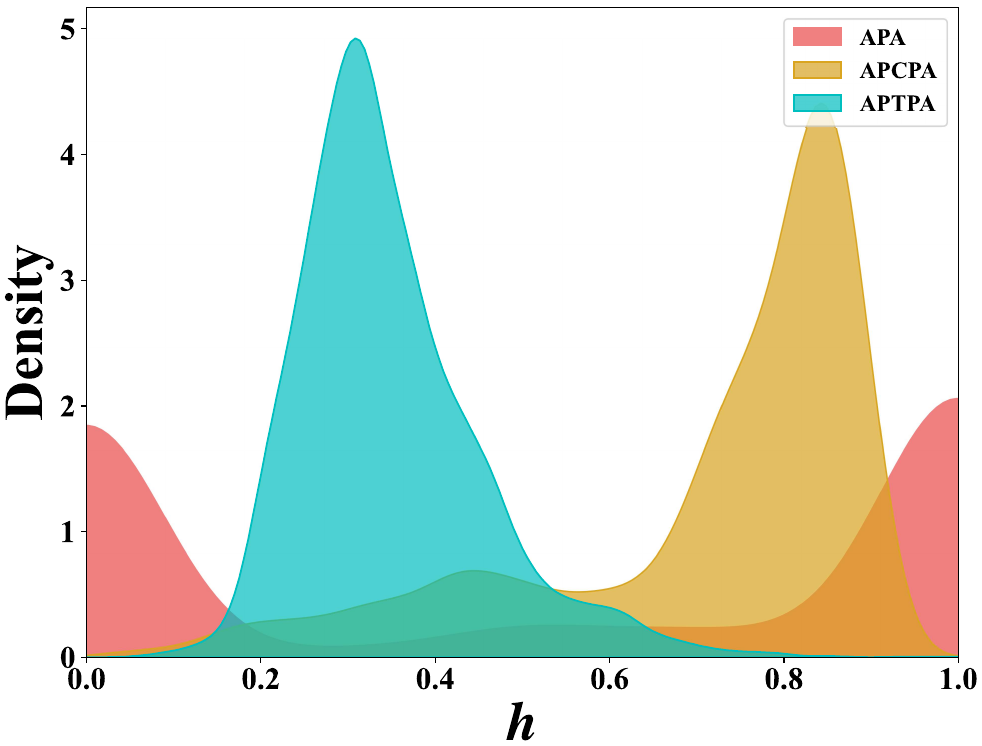}}
\hfil
\subfloat[Yelp (1D)\label{Yelp-1D}]{\includegraphics[width=0.3\linewidth]{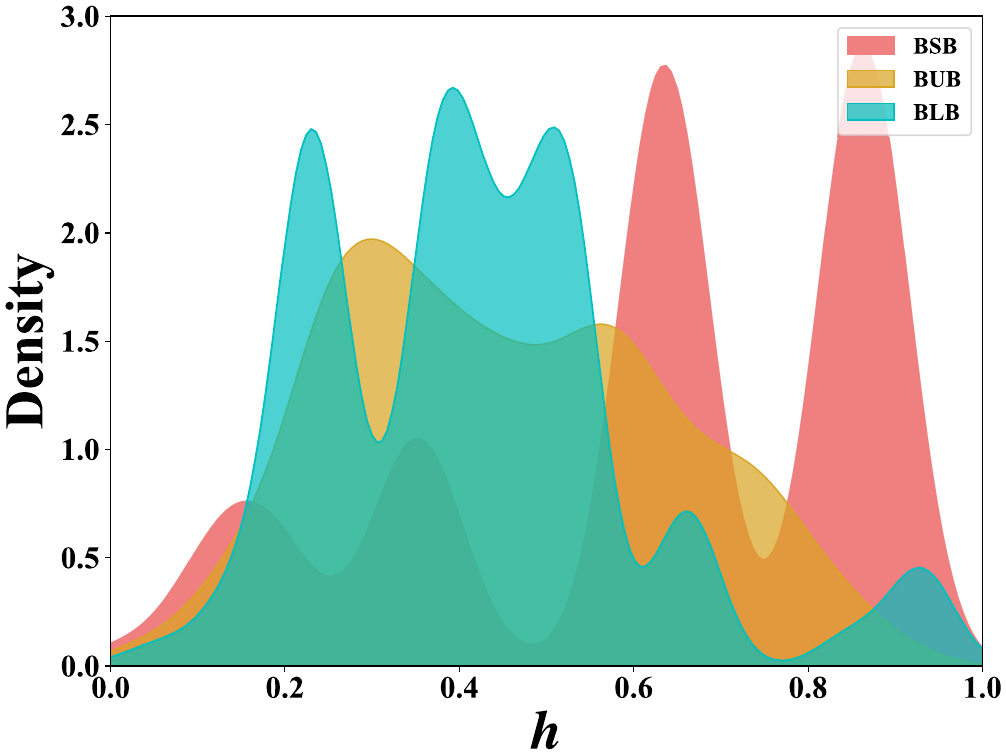}}

\subfloat[Yelp (BSB \& BUB)\label{SU}]{\includegraphics[width=0.3\linewidth]{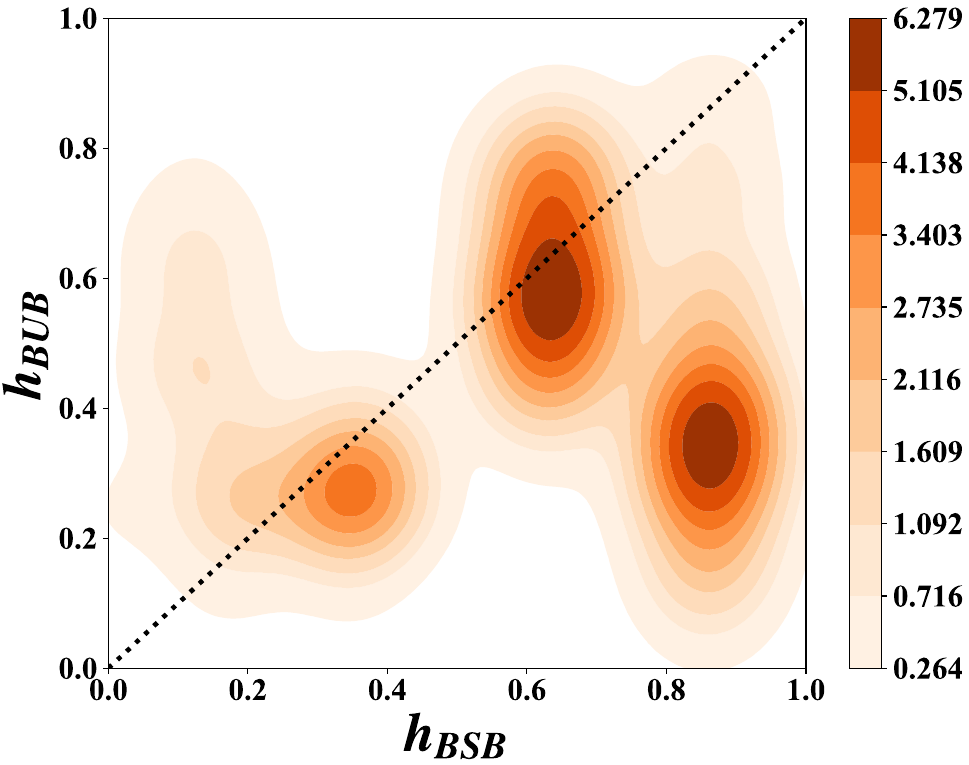}}
\hfil
\subfloat[Yelp (BSB \& BLB)\label{SL}]{\includegraphics[width=0.3\linewidth]{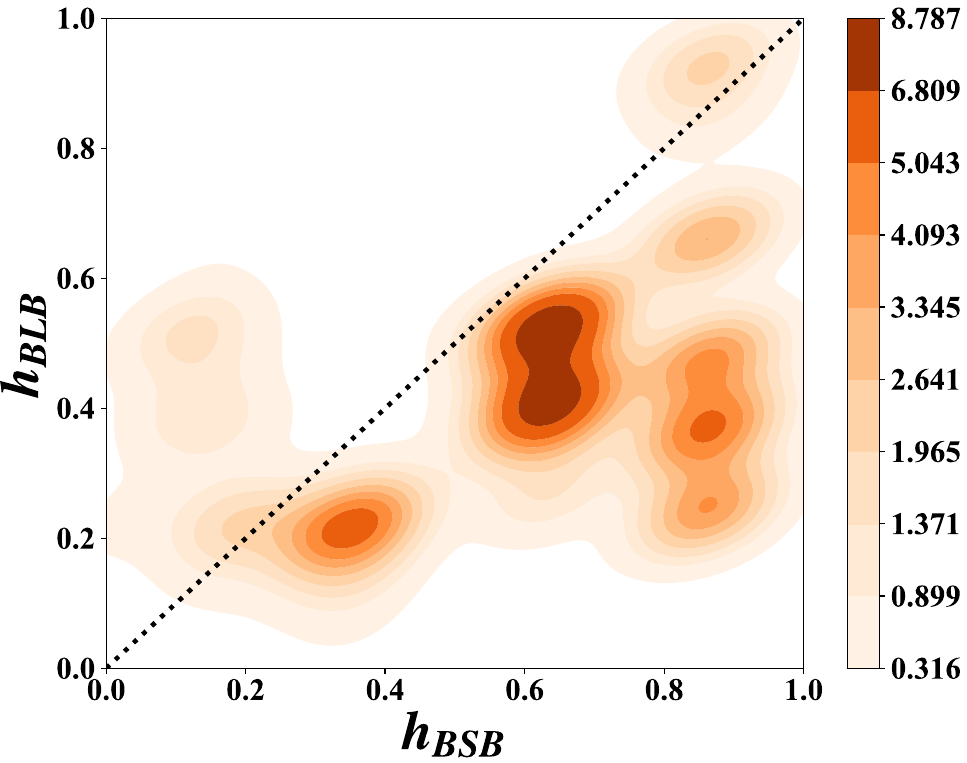}}
\hfil
\subfloat[Yelp (BUB \& BLB)\label{UL}]{\includegraphics[width=0.3\linewidth]{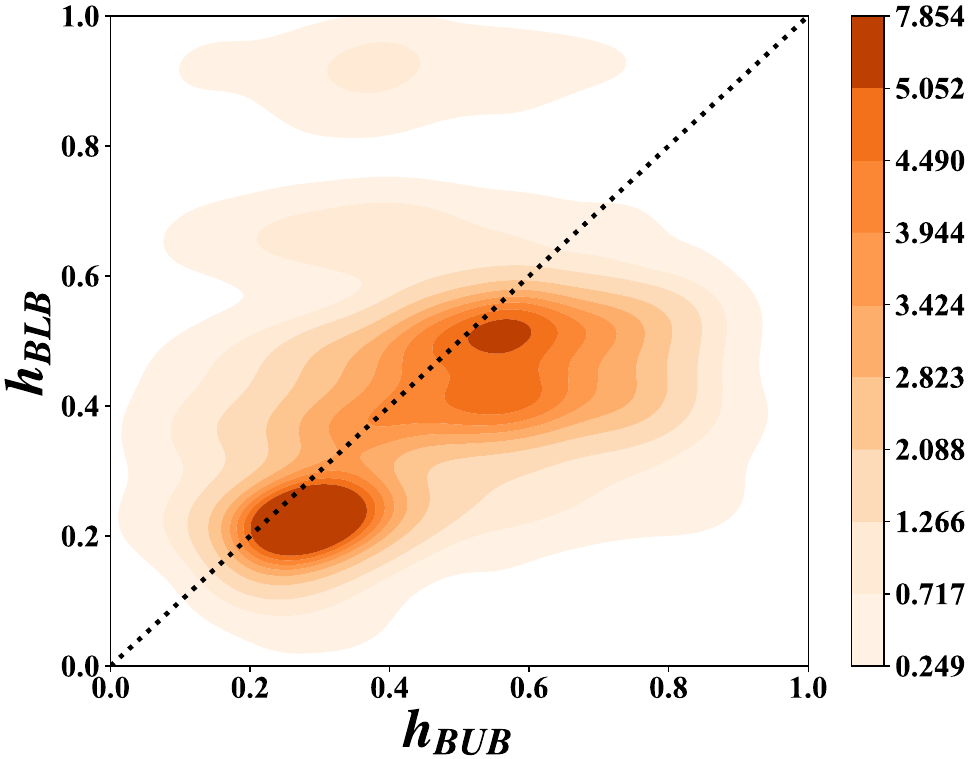}}
\caption{Node-level semantic homophily ratio distributions. Each real-world heterogeneous graph has diverse node neighborhood patterns.}
\label{NODE_SH}
\end{figure*}

\noindent \textbf{Definition 4. Node-level Semantic Homophily ratio (NHR).} Given a heterogeneous graph $\mathcal{G}$ with a meta-path set $\{\Phi_p\mid p=1,2,\cdots, P\}$, the node-level semantic homophily ratio is defined:

\begin{equation}
h(v_i)_{\Phi_p} = \frac{1}{\mid\mathcal{N}_i^{\Phi_p}\mid}\sum\limits_{v_j\in\mathcal{N}_i^{\Phi_p}}\mathbbm{1}(y_i=y_j)
\end{equation}
\begin{equation}
h(v_i) = \{h(v_i)_{\Phi_p}\mid p=1,2\cdots P\}
\end{equation}
where $\mathcal{N}_i^{\Phi_p}$ is the neighbors set of node $v_i$ based on meta-path $\Phi_p$ and $P$ is the number of meta-paths.

Fig. \ref{NODE_SH}\subref{ACM-1D}, \ref{NODE_SH}\subref{DBLP-1D}, and \ref{NODE_SH}\subref{Yelp-1D} present the Gaussian Kernel Density Estimation \cite{terrell1992variable} of NHR corresponding to each meta-path in the real-world datasets, and the other plots show the binary joint density estimation. In the univariate density plot, it can be observed that there is a significant distribution of nodes in both high and low homophily ratio regions across any given meta-path. For a meta-path with a higher MHR, there are more nodes in the high NHR region. In the binary joint density plot of the Yelp dataset, we could find that nodes with a high NHR based on BSB exhibit a varied NHR based on BUB. While some nodes display a high NHR, others show a low one. The same phenomenon could also be observed in other plots, implying that the neighborhood patterns of nodes in a heterogeneous graph are diverse. In conclusion, we summarize our empirical observation and identify three specific manifestations of semantic heterophily in heterogeneous graphs:

\begin{itemize}
\item \textbf{Observation 1:} For a given meta-path, the node-level semantic homophily ratio shows a diverse range, with some nodes having high ratios and others having low ratios.
\item \textbf{Observation 2:} Semantic homophily ratios at the node level based on various meta-paths exhibit diversity between distinct nodes, indicating that the relative ranking of semantic homophily ratios at the meta-path level cannot account for the neighborhood pattern of each node.
\item \textbf{Observation 3:} The reliability of the node features is not always guaranteed, as feature similarity may not always provide better category discriminability than graph structural proximity.
\end{itemize}

The three manifestations pose significant challenges in addressing UHGRL. In the context of semantic heterophily pointed out in Observations 1 and 2, the use of low-pass filtering could potentially result in a reduction in the distinctiveness of node representations for various categories within the local vicinity. Furthermore, the diverse patterns of node neighborhoods prevent the effective application of the semantic fusion mechanism with shared weights to all nodes. These limitations of the key components used in most existing methods hinder the generation of high-quality node representations. In addition, the intricate semantic heterophily inherent in structures and the limitations of node features pose formidable challenges in extracting positive samples. Mining of positive samples is of paramount importance in the context of the investigation of unsupervised representation learning \cite{velivckovic2018deep}. In contrast, the mere consideration of all meta-path-based neighbors as positive samples or the sole reliance on node features for positive sampling may lead to suboptimal performance.

\section{UHGRL under Semantic Heterophily}\label{LatGRL}

\begin{figure*}[!t]
	\centering
		\includegraphics[width=1.0\linewidth]{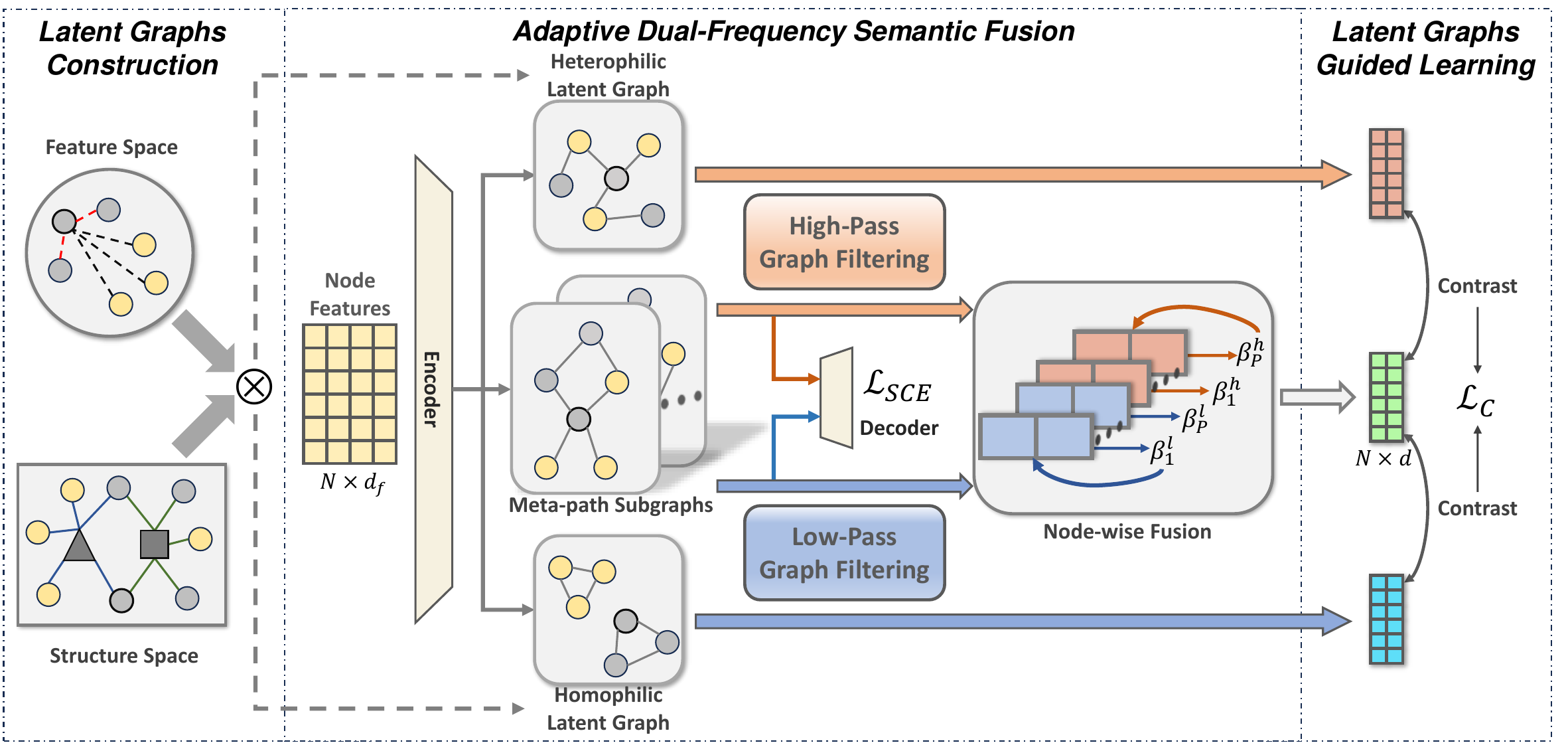}
	\caption{Illustration for our proposed framework LatGRL. It uses the coupled similarity measurement to construct a duo of latent graphs, guiding the representation learning. Additionally, it employs dual-pass graph filtering for node-wise adaptive fusion to tackle the challenges posed by semantic heterophily in heterogeneous graphs.}
	\label{Framework}
\end{figure*}

To overcome the above challenges, we propose Latent Graphs Guided Unsupervised Representation Learning (LatGRL). As illustrated in Fig. \ref{Framework}, LatGRL comprises three main modules: (1) \textit{construction of latent graphs} by coupling structure and feature similarity, (2) \textit{adaptive dual-frequency semantic fusion} using dual-pass graph filtering, and (3) \textit{latent graphs guided learning} that maximizes mutual information between the original semantically rich graph and latent graphs. Through the collaborative efforts of these three components, LatGRL performs adaptive semantic fusion and generates high-quality representations for nodes.

\subsection{Latent Graphs Construction}

Inspired by recent research on unsupervised homogeneous graph learning that leverages prior structures and features for positive sampling \cite{pan2023beyond}, we present a similarity mining approach that couples structures and features in heterogeneous graphs to achieve superior positive sample extraction. 

Two nodes sharing similar neighborhood structures often possess identical characteristics and labels \cite{perozzi2014deepwalk}.
Hence, we begin by focusing on structure similarity and propose a simple yet effective methodology. Within the heterogeneous graphs, gauging the structure similarity amongst nodes necessitates a comprehensive evaluation of their topological structures spanning various relations. Based on this, we propose the concept of global structure similarity. Formally, the similarity of the global structure between two nodes of identical type is evaluated by considering all meta-path subgraphs $\{\mathcal{G}_{\Phi_p}\mid p=1,2,\cdots, P\}$:

\begin{equation}
\mathbf{M}^{\Phi_p} = \tilde{\mathbf{D}}_{\Phi_p}\tilde{\mathbf{A}}^{\Phi_p},\quad \mathbf{M} = \frac{1}{P}\sum\limits_{p=1}^P\mathbf{M}^{\Phi_p}
\end{equation}
\begin{equation}
sim^T(v_i, v_j) = \frac{\mathbf{M}_{i\cdot}^{\top}\mathbf{M}_{j\cdot}}{\parallel\mathbf{M}_{i\cdot}\parallel\cdot\parallel\mathbf{M}_{j\cdot}\parallel} = \left\langle\mathbf{M}_{i\cdot},\mathbf{M}_{j\cdot}\right\rangle
\end{equation}
where $\mathbf{M}^{\Phi_p}$ represents random walk normalized adjacency matrix of meta-path subgraph $\mathcal{G}_{\Phi_p}$. We obtain the probability diffusion matrix $\mathbf{M}$ by averaging over all meta-paths. The element of $\mathbf{M}$ represents the combined probability of each node randomly walking to other nodes through multiple relations, capturing its local structural information. The structure similarity between two nodes is defined as the cosine value of their diffusion vectors. A larger value indicates a higher proportion of shared neighbors along all meta-paths.

Subsequently, we achieve the ultimate measurement of node similarity within the heterogeneous graph by adaptively amalgamating both global structures and features. Specifically, the similarity between the nodes $v_i$ and $v_j$ can be represented as: $sim(v_i, v_j) = sim^T(v_i, v_j)\cdot sim^F(v_i, v_j)$, where $sim^F(v_i, v_j)=\frac{\mathbf{X}_{i\cdot}^{\top}\mathbf{X}_{j\cdot}}{\parallel\mathbf{X}_{i\cdot}\parallel\cdot\parallel\mathbf{X}_{j\cdot}\parallel}$ denotes the cosine similarity of node features. It should be noted that the initial node features have undergone normalization, ensuring that every element adheres to a range greater than or equal to zero. Consequently, the calculated similarity values $sim(v_i, v_j)\in[0,1]$.

To overcome the semantic heterophily inherent in structures, we embark on constructing a duo latent graph by using the coupled similarity measurement: the homophilic latent graph and the heterophilic latent graph. Specific formulas are as follows:

% \begin{equation}
% \mathbf{S}^T_{i,j} = \frac{\mathbf{M}_{i\cdot}^{\top}\mathbf{M}_{j\cdot}}{\parallel\mathbf{M}_{i\cdot}\parallel\cdot\parallel\mathbf{M}_{j\cdot}\parallel},\mathbf{S}^F_{i,j} = \frac{\mathbf{X}_{i\cdot}^{\top}\mathbf{X}_{j\cdot}}{\parallel\mathbf{X}_{i\cdot}\parallel\cdot\parallel\mathbf{X}_{j\cdot}\parallel},
% \end{equation}
\begin{equation}
\mathbf{S}^T_{i,j} = sim^T(v_i, v_j), \quad \mathbf{S}^F_{i,j} = sim^F(v_i, v_j)
\end{equation}
\begin{equation}
\mathbf{S} = \mathbf{S}^T\odot \mathbf{S}^F
\end{equation}
\begin{equation}
\mathbf{W} = (1. - \mathbf{S}^T)\odot (1. -\mathbf{S}^F)
\end{equation}
\begin{equation}\label{latent_graphs_cons}
\mathbf{A}^S = \mathop\mathrm{Top}(\mathbf{S} - \mathbf{I}, K),\quad \mathbf{A}^W = \mathop\mathrm{Top}(\mathbf{W}, K)
\end{equation}
where $\odot$ denotes the Hadamard product. $\mathbf{S}$ represents the similarity matrix, where a higher value indicates a greater probability of belonging to the same category. On the contrary, $\mathbf{W}$ represents the dissimilarity matrix. $\mathop\mathrm{Top}(\cdot, K)$ refers to selecting the top $K$ elements of each row, setting their values to 1, and setting the other values to 0. $\mathbf{A}^S$ and $\mathbf{A}^W$ can be seen as adjacency matrices of homophilic and heterophilic latent graphs, respectively. In fact, for each node, when constructing the homophilic latent graph, we consider searching for the most similar nodes among the 1-hop and 2-hop meta-path-based neighbors of the same type, excluding the node itself. On the other hand, the heterophilic latent graph focuses more on the extraction of negative samples that are distant in the topological structure.

\subsection{Adaptive Dual-frequency Semantic Fusion}

Due to the numerous heterophilic connections, we introduce high-pass graph filtering to sharpen the features of neighboring nodes, thereby avoiding the confusion of representations from different categories. Learning complex filters often requires supervision signals from label information, which is not suitable for unsupervised tasks \cite{chen2023polygcl}. In contrast, we propose a simple and interpretable mechanism. We use $\tilde{\mathbf{A}}^{\Phi}_{sym}$ as the low-pass filter to capture low-frequency information and preserve commonalities among neighborhoods. Simultaneously, $\tilde{\mathbf{L}}^{\Phi}_{sym}$ is employed as the high-pass filter to retain high-frequency information and sharpen the representations of nodes along the edges. Graph filtering is applied on the node embedding:

\begin{equation}
\mathbf{H}^{\Phi_p, l} = (\tilde{\mathbf{A}}^{\Phi_p}_{sym})^rf(\mathbf{X})
\end{equation}
\begin{equation}
\mathbf{H}^{\Phi_p, h} = (\tilde{\mathbf{L}}^{\Phi_p}_{sym})^rf(\mathbf{X})
\end{equation}
\begin{equation}\label{dual_frequency_rep}
\tilde{\mathbf{H}}^{\Phi_p, l}_{i\cdot} = \text{norm}(\mathbf{H}^{\Phi_p, l}_{i\cdot}),  \quad \tilde{\mathbf{H}}^{\Phi_p, h}_{i\cdot} = \text{norm}(\mathbf{H}^{\Phi_p, h}_{i\cdot})
\end{equation}
% \begin{equation}
% \tilde{\mathbf{h}}_{i,{\Phi}_p}^l = norm(\mathbf{h}_{i,{\Phi}_p}^l),  \quad \tilde{\mathbf{h}}_{i,{\Phi}_p}^h = norm(\mathbf{h}_{i,{\Phi}_p}^h)
% \end{equation}
where $f(\cdot):\mathbbm{R}^{d_f}\rightarrow\mathbbm{R}^{d}$ denotes the node encoder, which is implemented using MLP in this study. $\mathbf{H}^{\Phi_p, l}$ and $\mathbf{H}^{\Phi_p, h}$ are the low-frequency smoothed representations and the high-frequency sharpened representations in the context of meta-path $\Phi_p$, respectively. $r$ is the filter order. To eliminate the influence of representation magnitude, we apply L2 Normalization to each node representation, given as $\text{norm}(\mathbf{h})=\frac{\mathbf{h}}{\parallel\mathbf{h}\parallel}$.

Subsequently, the two representations of each meta-path view are collectively fed into a shared decoder $g(\cdot):\mathbbm{R}^{d}\rightarrow\mathbbm{R}^{d_f}$. The Scaled Cosine Error (SCE) \cite{hou2022graphmae} is employed as the reconstruction loss to ensure that the representations of each meta-path preserve sufficient information. The contribution of simple samples can be controlled by adjusting the sharpening parameter $\gamma$:

\begin{equation}
\hat{\mathbf{X}}^p = g([\tilde{\mathbf{H}}^{\Phi_p, l} \parallel \tilde{\mathbf{H}}^{\Phi_p, h}])
\end{equation}
\begin{equation}\label{reconstruction_loss}
\mathcal{L}_{SCE} =  \frac{1}{N\cdot P}\sum\limits_{p=1}^P \sum\limits_{i=1}^{N}\left(1 - \frac{\mathbf{X}_{i\cdot}^{\top} \hat{\mathbf{X}}_{i\cdot}^p}{\parallel \mathbf{X}_{i\cdot}\parallel\cdot \parallel \hat{\mathbf{X}}_{i\cdot}^p\parallel} \right)^{\gamma}
\end{equation}
where $[\cdot\parallel\cdot]$ denotes the row-wise concatenation operation. The decoder is also implemented using an MLP in this study.

Section \ref{4} emphasizes that the node-level semantic homophily ratios based on various meta-paths exhibit diversity across distinct nodes. This finding underscores the necessity of employing an adaptive encoding mechanism to handle the varied patterns of node neighborhoods. Therefore, we propose an innovative approach, dubbed the adaptive dual-frequency semantic fusion method for node-wise modeling:

\begin{equation}
\omega_{i,p}^l=\sigma(\mathbf{q}_l^{\top} \tilde{\mathbf{H}}^{\Phi_p, l}_{i\cdot}), \quad \omega_{i,p}^h=\sigma(\mathbf{q}_h^{\top} \tilde{\mathbf{H}}^{\Phi_p, h}_{i\cdot})
\end{equation}
\begin{equation}
\beta_{i,p}^l=\frac{exp(\omega_{i,p}^l)}{\sum_{j=1}^P exp(\omega_{i,j}^l)+\sum_{j=1}^P exp(\omega_{i,j}^h)}
\end{equation}
\begin{equation}
\beta_{i,p}^h=\frac{exp(\omega_{i,p}^h)}{\sum_{j=1}^P exp(\omega_{i,j}^l)+\sum_{j=1}^P exp(\omega_{i,j}^h)}
\end{equation}
\begin{equation}\label{semantic_fusion_rep}
\mathbf{Z}_{i}=\sum\limits_{p=1}^P\beta_{i,p}^l \cdot \tilde{\mathbf{H}}^{\Phi_p, l}_{i\cdot} + \sum\limits_{p=1}^P\beta_{i,p}^h \cdot \tilde{\mathbf{H}}^{\Phi_p, h}_{i\cdot}
\end{equation}
where $\mathbf{q}_l$ and $\mathbf{q}_h$ denote the learnable attention vectors associated with low-frequency and high-frequency information, respectively. $\beta_{i,p}^l$ and $\beta_{i,p}^h$ are the fusion weights for the low-pass and high-pass representations of node $v_i$ in the meta-path $\Phi_p$.
$\sigma(\cdot)$ is the non-linear activation function. $\mathbf{Z}_{i}$ represents the final representation of node $v_i$, which would be used in the downstream tasks. Many existing UHGRL methods use shared semantic fusion weights for all nodes \cite{wang2021self,yu2023heterogeneous}. On the contrary, our method focuses on adaptive dual-frequency fusion, tailored for nodes with different neighborhood patterns.

\subsection{Latent Graphs Guided Learning}

We address the lack of label information in UHGRL by exploring self-supervised signals from the data. Unlike existing methods that rely on predefined data augmentation techniques like random edge deletion or random feature removal to generate contrastive views \cite{you2020graph}, which could be harmful in heterophily graphs \cite{zhu2020beyond}, our approach leverages latent graphs to provide valuable supervision signals. The homophilic latent graph mainly encompasses nodes with shared class neighbors, so we use low-pass filtering to extract common traits within the same category. In contrast, the heterophilic latent graph comprises nodes whose neighbors represent different categories, motivating high-pass filtering to reveal differentiating information among distinct categories. The process of latent graph encoding is as follows:

\begin{equation}\label{latent_graph_rep}
\mathbf{Z}^{l} = (\tilde{\mathbf{A}}^{S}_{sym})^rf(\mathbf{X}), \quad
\mathbf{Z}^{h} = (\tilde{\mathbf{L}}^{W}_{sym})^rf(\mathbf{X})
\end{equation}
where $\tilde{\mathbf{A}}^{S}_{sym}$ and $\tilde{\mathbf{L}}^{W}_{sym}$ are the renormalized versions. Then, we maximize the mutual information between the semantic fusion representations $\mathbf{Z}$ and the latent graph representations $\mathbf{Z}^l$ and $\mathbf{Z}^h$, where $\mathbf{Z}^l$ provides communal information within the same category and $\mathbf{Z}^h$ captures distinction characteristics among different categories. Both of them collectively guide the representation learning process. The optimization objective can be written as follows:

\begin{equation}
\mathop{\max}_{\theta} I(\mathbf{Z}, \mathbf{Z}^l) + I(\mathbf{Z}, \mathbf{Z}^h)
\end{equation}
where $\theta$ is the parameters of the model. The InfoNCE objective is used as the lower bound of mutual information \cite{he2020momentum}. We also employ commonly used positive sampling techniques in contrastive learning. Unlike previous methods that relied on neighboring nodes \cite{wang2021self} or employed the KNN algorithm \cite{liu2023beyond}, we leverage the previously defined similarity measurement coupling global structures and features to select positive samples. Specifically, for a given node $v_i$, the positive samples set $\mathbbm{P}_i = \mathop{\mathrm{\arg top}}_{j} (sim(v_i, v_j), k_{pos})$, which means selecting the top $k_{pos}$ nodes that have the highest similarity with $v_i$ as its positive samples. $\mathbf{Z}$, $\mathbf{Z}^l$, and $\mathbf{Z}^h$ are projected to a shared latent space using separate learnable MLPs for fair similarity measurement and loss calculation. The contrastive loss for node $v_i$ can be formulated as follows:

\begin{equation}\label{CONTRASTIVE-LOSS}
\mathcal{L}(\mathbf{Z}_i, \mathbf{Z}^l_i) = -\log\frac{\sum_{v_j\in \mathbbm{P}_i}e^{\left\langle \tilde{\mathbf{Z}}_i,\ \tilde{\mathbf{Z}}^l_j\right\rangle  / \tau}}{\sum_{v_k\in \mathcal{V}_t} e^{\left\langle \tilde{\mathbf{Z}}_i,\ \tilde{\mathbf{Z}}^l_k \right\rangle  / \tau}}
\end{equation}
where $\tilde{\mathbf{Z}}_i$ is non-linear projection of $v_i$'s representation in the latent space, $\left\langle \cdot , \cdot\right\rangle$ refers to the cosine similarity, and $\tau$ is the temperature parameter.
During full graph training, $\mathcal{V}_t$ represents the entire set of target nodes. However, for the mini-batch training, $\mathcal{V}_t$ denotes the sampled nodes set within each batch.
The contrastive loss is defined as the average of all target nodes:

\begin{equation}\label{contrastive_loss}
\begin{aligned}
\mathcal{L}_{C} = &\frac{1}{2N} \sum\limits_{i=1}^N [\mathcal{L}(\mathbf{Z}_i, \mathbf{Z}^l_i)+\mathcal{L}(\mathbf{Z}_i^l, \mathbf{Z}_i)] + \\&
\frac{1}{2N}\sum\limits_{i=1}^N [\mathcal{L}(\mathbf{Z}_i, \mathbf{Z}^h_i)+\mathcal{L}(\mathbf{Z}_i^h, \mathbf{Z}_i)]
\end{aligned}
\end{equation}

Consequently, the overall objective of LatGRL, which we aim to minimize, consists of two loss terms:

\begin{equation}
\mathcal{L} = \mathcal{L}_{C} + \mathcal{L}_{SCE}
\end{equation}

We provide the overall process of LatGRL in the Algorithm \ref{LatGRL_algorithm}. The concurrent guidance from homophilic and heterophilic latent graphs offers valuable supervision signals for unsupervised representation learning, while also seamlessly accommodating single meta-paths or homogeneous graphs. The mechanism of latent graphs-guided learning remains unexplored in the current landscape of graph learning, highlighting the innovation of our approach in both heterogeneous and homogeneous graph domains.

\normalem

\begin{algorithm}[t]
  \SetAlgoLined
  \KwIn{Heterogeneous graph $\mathcal{G}$; Node attributes $\mathbf{X}$; Meta-path set $\{\Phi_1, \cdots, \Phi_P\}$; Epochs $E$} 
  \KwOut{Learned node representations $\mathbf{Z}$} 

  Initialize parameters; \\
  Construct latent graphs $\{\mathbf{A}^S,\mathbf{A}^W\}$ by Eq.(\ref{latent_graphs_cons});\\
\For{e=1,2,...,E}{
  \For{each $\Phi_p$ in $\{\Phi_1, \cdots, \Phi_P\}$}{ 
  Obtain dual-frequency node representations $\{\tilde{\mathbf{H}}^{\Phi_p, l}, \tilde{\mathbf{H}}^{\Phi_p, h}\}$ by Eq.(\ref{dual_frequency_rep}); \\
  }
  Calculate the reconstruction loss $\mathcal{L}_{SCE}$ by Eq.(\ref{reconstruction_loss}); \\
    Obtain fusion node representations $\mathbf{Z}$ by Eq.(\ref{semantic_fusion_rep}); \\
    Obtain latent graphs node representations $\{\mathbf{Z}^l, \mathbf{Z}^h\}$ by Eq.(\ref{latent_graph_rep}); \\
    Calculate the contrastive loss $\mathcal{L}_C$ by Eq.(\ref{contrastive_loss}); \\
    Calculate the total loss $\mathcal{L}$ and update parameters; \\
  }
  return node representations $\mathbf{Z}$;
  \caption{The optimization of LatGRL}
    \label{LatGRL_algorithm}
\end{algorithm}

\ULforem

\subsection{Scalable Implementation}

To accommodate large-scale heterogeneous graph data, we develop a scalable implementation called LatGRL-S, which consists of the following two steps.

\textbf{Scalable Latent Graph Construction.} In the process of latent graph construction, the calculation of the similarity between each node and all other nodes is impractical for real-world graphs. Therefore, we only use first-order neighboring nodes to construct the homophilic latent graph, as the number of first-order neighbors in large-scale graphs is typically substantial. In contrast, for the heterophilic latent graph, we employ an anchor-based construction approach. Formally:

\begin{equation}
\mathcal{N}^S_i =  \mathop{\mathrm{\arg top}}_{v_j\in \bar{\mathcal{N}}_i} \left (sim^T(v_i, v_j)\cdot sim^F(v_i, v_j), K\right )
\end{equation}
\begin{equation}
\mathcal{N}^W_i =  \mathop{\mathrm{\arg top}}_{v_j\in U} \left ([1-sim^T(v_i, v_j)]\cdot [1-sim^F(v_i, v_j)], K\right )
\end{equation}
where $\bar{\mathcal{N}}_i = \bigcup_{p=1}^{P}\mathcal{N}_{\Phi_p}$ represents all first-order neighbors of node $v_i$ based on all meta-paths and $U$ is the anchor set consisting of $m$ randomly selected nodes. $\mathcal{N}^S_i$ and $\mathcal{N}^W_i$ denote the neighbor sets of node $v_i$ in the homophilic and heterophilic latent graph, respectively. It is worth noting that we are not performing simple neighbor sampling. The homophilic latent graph primarily focuses on local information, constructing personalized homophilic neighborhoods for each node based on its local context. For the heterophilic latent graph, the anchor graph construction could emphasize global information to capture inter-class differences.

\textbf{Mini-Batch Training with Pre-Filtering.} In the implementation of LatGRL, graph filtering is performed on the embeddings, resulting in filtering calculations being executed during each training epoch. To enhance training efficiency, we first pre-filter the raw features to obtain low-frequency and high-frequency features before training. Subsequently, during the mini-batch training, the two features are fed to the encoding layer. The entire process is illustrated as follows:

\textit{Pre-Filtering:}

\begin{equation}
\begin{aligned}
\mathbf{X}^{{\Phi}_p,l}=(\tilde{\mathbf{A}}^{\Phi_p}_{sym})^r\mathbf{X}, \quad \mathbf{X}^{{\Phi}_p,h}=(\tilde{\mathbf{L}}^{\Phi_p}_{sym})^r\mathbf{X}
\end{aligned}
\end{equation}

\textit{Mini-Batch Training:}

\begin{equation}
\mathbf{H}^{\Phi_p, l}_B = f(\mathbf{X}^{{\Phi}_p,l}_B), \quad
\mathbf{H}^{\Phi_p, h}_B = f(\mathbf{X}^{{\Phi}_p,h}_B)
\end{equation}
\begin{equation}
\mathbf{Z}_B = \text{Fusion}(\mathbf{H}^{\Phi_1, l}_B,\cdots, \mathbf{H}^{\Phi_P, l}_B; \ \mathbf{H}^{\Phi_1, h}_B,\cdots, \mathbf{H}^{\Phi_P, h}_B)
\end{equation}
where $\mathbf{H}^{\Phi_p, l}_B$ and $\mathbf{H}^{\Phi_p, h}_B$ represents the low-frequency and high-frequency representations of the sampled nodes in each batch under the meta-path $\Phi_p$, respectively. $\mathbf{Z}_B$ denotes the final representations of these nodes. The node representations of latent graphs are also obtained through pre-filtering and dimensionality reduction processes. This decoupled implementation approach enhances training efficiency by eliminating the need for time-consuming and resource-intensive neighbor sampling and message aggregation operations in each mini-batch. Following LatGRL, we then employ the latent graphs guided learning method to achieve model training. The loss is computed only for the nodes within the batch.

\subsection{Time Complexity}

Due to the sparsity of real-world heterogeneous graphs, we implement graph filtering with sparse matrix techniques. For simplicity, we use the same $D$ as the dimensionality of input features and node representations, and $B$ as the batch size of mini-batch training and the number of anchors for scalable latent graph construction. $N$ and $E$ are the number of target nodes and edges. $r$ is the filter order. Since latent graph construction only needs to be done once before training, we divide the analysis into two stages: Preprocessing and Training. For LatGRL, the complexity of the preprocessing stage is $\mathcal{O}\left(N^2(N+D)\right)$. The training stage includes graph filtering and loss calculation, with a complexity of $\mathcal{O}\left(D(N^2+rE+ND)\right)$. For LatGRL-S, the preprocessing stage involves scalable latent graph construction and pre-filtering, with a complexity of $\mathcal{O}\left(E(N+rD)+NB(N+D)\right)$. The training stage has a complexity of $\mathcal{O}\left(ND(B+D)\right)$. We compare our complexity with baselines in Table \ref{time_complexity}. It can be observed that existing methods require a complexity of at least $\mathcal{O}(N^2)$ during the training phase. In contrast, our method is only linear, demonstrating the significant advantage of LatGRL-S in handling large-scale graph data.

% % Table generated by Excel2LaTeX from sheet 'Sheet1'
% \begin{table}[htbp]
% \centering
% \caption{The time complexity analysis.}
% \resizebox{\linewidth}{!}
% {
%     \begin{tabular}{ccccc}
%     \toprule
%     \multirow{2}[4]{*}{Method} & \multirow{2}[4]{*}{HeCo} & \multirow{2}[4]{*}{HGMAE} & \multicolumn{2}{c}{LatGRL-S} \\
% \cmidrule{4-5}          &       &       & Preprocessing & Training \\
%     \midrule
%     Complexity & $\mathcal{O}\left(D(N^2+E+ND)\right)$     & $\mathcal{O}\left(N^2+D(E+ND)\right)$     & $\mathcal{O}\left(E(N+rD)+NB(N+D)\right)$     & $\mathcal{O}\left(ND(B+D)\right)$ \\
%     \bottomrule
%     \end{tabular}%
%     }
%   \label{time_complexity}%
% \end{table}%

\begin{table}[t]
\centering
\caption{The time complexity in every training epoch.}
\resizebox{\linewidth}{!}
{
\renewcommand{\arraystretch}{1.25}
    \begin{tabular}{cccc}
    \toprule
    Method & HeCo  & HGMAE & LatGRL-S \\
    \midrule
    Complexity & $\mathcal{O}\left(D(N^2+E+ND)\right)$     & $\mathcal{O}\left(N^2+D(E+ND)\right)$    & $\mathcal{O}\left(ND(B+D)\right)$ \\
    \bottomrule
    \end{tabular}%
    }
  \label{time_complexity}%
\end{table}%

\section{Experiments}
\subsection{Experimental setup}

\textbf{Datasets.} We employ four benchmark heterogeneous attributed graph datasets and a large-scale heterogeneous graph dataset. The statistics of these datasets are presented in Table \ref{DATASETS}.
\begin{itemize}
\item \textbf{DBLP}\cite{zhao2020network}. DBLP is extracted from the computer science bibliography website and the target nodes are authors that are divided into 4 categories.
\item \textbf{ACM}\cite{fu2020magnn}. ACM is an academic paper dataset and the target nodes are papers that are divided into 3 categories.
\item \textbf{IMDB}\cite{fu2020magnn}. IMDB is a subset of the Internet Movie Database and the target nodes are movies that are divided into 3 categories.
\item \textbf{Yelp}\cite{shi2020rhine}. Yelp is extracted from the merchant review website and the target nodes are businesses that are divided into 3 categories.
\item \textbf{Ogbn-mag}\cite{wang2020microsoft}. Ogbn-mag is a subset of the Microsoft Academic Graph and the target nodes are papers that are divided into 349 categories.
\end{itemize}

\textbf{Baselines.} We compare LatGRL with 11 other state-of-the-art methods, which can be grouped into three categories:
\begin{itemize}
\item \textbf{Semi-supervised Learning Methods:} GAT\cite{velivckovic2018graph} incorporates neighborhood attention mechanisms for homogeneous graphs, while HAN\cite{wang2019heterogeneous} uses hierarchical attention for heterogeneous graphs.

\item \textbf{Unsupervised Heterogeneous Graph Representation Learning Methods:} DGI\cite{velivckovic2018deep} maximizes agreement between node representations and global representations. GraphMAE\cite{hou2022graphmae} is a masked graph autoencoder that incorporates graph masking and feature reconstruction. DGI and GraphMAE are both methods for homogeneous graphs.
Mp2vec\cite{dong2017metapath2vec} is a shallow heterogeneous graph embedding method using meta-path-based random walks. DMGI\cite{park2020unsupervised} maximizes the mutual information between local and global information and includes semantic consistency constraints. HeCo\cite{wang2021self} constructs contrastive loss between meta-paths and network schemas. MEOW\cite{yu2023heterogeneous} contrasts coarse-grained and fine-grained views and incorporates negative sample importance mining. DuaLGR\cite{ling2023dual} is a multi-view graph clustering method that introduces structural and feature pseudo labels. HGMAE\cite{tian2023heterogeneous} explores masked autoencoders in heterogeneous graphs.
\item \textbf{Unsupervised Learning Methods For Heterophily:} GREET\cite{liu2023beyond} addresses neighborhood heterophily in homogeneous graphs by using an edge discriminator to differentiate between homophilic and heterophilic edges.
\end{itemize}

\begin{table}[t]
\centering
\caption{The statistics of the datasets.}
\resizebox{\linewidth}{!}
{
\renewcommand{\arraystretch}{1.25}
\begin{tabular}{|c|c|c|c|c|}
\hline
Datasets                  & Node                                                                                                                                        & Relation                                                                                                           & Meta-path                                                                    & Features              \\ \hline
\multirow{4}{*}{DBLP}     & \multirow{4}{*}{\begin{tabular}[c]{@{}c@{}}Author (A): 4057\\ Paper (P): 14328\\ Conference (C): 20\\ Term (T): 7723\end{tabular}}          & \multirow{4}{*}{\begin{tabular}[c]{@{}c@{}}P-A: 19645\\ P-C: 14328\\ P-T: 85810\end{tabular}}                      & \multirow{4}{*}{\begin{tabular}[c]{@{}c@{}}APA\\ APCPA\\ APTPA\end{tabular}} & \multirow{4}{*}{334}  \\
                          &                                                                                                                                             &                                                                                                                    &                                                                              &                       \\
                          &                                                                                                                                             &                                                                                                                    &                                                                              &                       \\
                          &                                                                                                                                             &                                                                                                                    &                                                                              &                       \\ \hline
\multirow{3}{*}{ACM}      & \multirow{3}{*}{\begin{tabular}[c]{@{}c@{}}Paper (P): 4019\\ Author (A): 7167\\ Subject (S): 60\end{tabular}}                               & \multirow{3}{*}{\begin{tabular}[c]{@{}c@{}}P-A: 13407\\ P-S: 4019\end{tabular}}                                    & \multirow{3}{*}{\begin{tabular}[c]{@{}c@{}}PAP\\ PSP\end{tabular}}           & \multirow{3}{*}{1902} \\
                          &                                                                                                                                             &                                                                                                                    &                                                                              &                       \\
                          &                                                                                                                                             &                                                                                                                    &                                                                              &                       \\ \hline
\multirow{3}{*}{IMDB}     & \multirow{3}{*}{\begin{tabular}[c]{@{}c@{}}Movie (M): 4278\\ Director (D): 2081\\ Actor (A): 5257\end{tabular}}                             & \multirow{3}{*}{\begin{tabular}[c]{@{}c@{}}M-D: 4278\\ M-A: 12828\end{tabular}}                                    & \multirow{3}{*}{\begin{tabular}[c]{@{}c@{}}MDM\\ MAM\end{tabular}}           & \multirow{3}{*}{3066} \\
                          &                                                                                                                                             &                                                                                                                    &                                                                              &                       \\
                          &                                                                                                                                             &                                                                                                                    &                                                                              &                       \\ \hline
\multirow{4}{*}{Yelp}     & \multirow{4}{*}{\begin{tabular}[c]{@{}c@{}}Business (B): 2614\\ User (U): 1286\\ Service (S): 4\\ Rating Levels (L): 9\end{tabular}}        & \multirow{4}{*}{\begin{tabular}[c]{@{}c@{}}B-U: 30838\\ B-S: 2614\\ B-L: 2614\end{tabular}}                        & \multirow{4}{*}{\begin{tabular}[c]{@{}c@{}}BUB\\ BSB\\ BLB\end{tabular}}     & \multirow{4}{*}{82}   \\
                          &                                                                                                                                             &                                                                                                                    &                                                                              &                       \\
                          &                                                                                                                                             &                                                                                                                    &                                                                              &                       \\
                          &                                                                                                                                             &                                                                                                                    &                                                                              &                       \\ \hline
\multirow{4}{*}{Ogbn-mag} & \multirow{4}{*}{\begin{tabular}[c]{@{}c@{}}Paper (P): 736389\\ Author (A): 1134649\\ Institution (I): 8740\\ Field (F): 59965\end{tabular}} & \multirow{4}{*}{\begin{tabular}[c]{@{}c@{}}P-A: 7145660\\ P-P: 5416271\\ P-F: 7505078\\ A-I: 1043998\end{tabular}} & \multirow{4}{*}{\begin{tabular}[c]{@{}c@{}}PP\\ PAP\end{tabular}}            & \multirow{4}{*}{128}  \\
                          &                                                                                                                                             &                                                                                                                    &                                                                              &                       \\
                          &                                                                                                                                             &                                                                                                                    &                                                                              &                       \\
                          &                                                                                                                                             &                                                                                                                    &                                                                              &                       \\ \hline
\end{tabular}
}
\label{DATASETS}
\end{table}

\begin{table*}[htbp]
  \centering
  \caption{Quantitative results ($\%$±$\sigma$) on node classification. The best, runner-up and third-best results are highlighted using \textbf{bold}, \uline{underline}, and \uwave{under-wave}, respectively.}
  \resizebox{\textwidth}{!}
  {
    \begin{tabular}{c|c|c|cc|cccccccc|c|c|c}
    \toprule
    \multirow{2}[2]{*}{Datasets} & \multirow{2}[2]{*}{Metric} & \multirow{2}[2]{*}{Split} & GAT   & HAN   & Mp2vec & DGI   & DMGI  & HeCo  & GraphMAE & MEOW  & DuaLGR & HGMAE & GREET & \multirow{2}[2]{*}{\textbf{LatGRL-S}} & \multirow{2}[2]{*}{\textbf{LatGRL}} \\
          &       &       & 2018  & 2019  & 2017  & 2019  & 2020  & 2021  & 2022  & 2023  & 2023  & 2023  & 2023  &       &  \\
    \midrule
    \multirow{9}[6]{*}{DBLP} & \multirow{3}[2]{*}{Ma-F1} & 20    & 90.60±0.4 & 89.31±0.9 & 88.98±0.2 & 87.93±2.4 & 89.94±0.4 & 91.28±0.2 & 88.76±0.3 & \uwave{92.57±0.4} & 90.56±0.3 & 92.28±0.5 & 78.44±0.9 & \uline{92.65±0.4} & \textbf{94.23±0.2} \\
          &       & 40    & 90.72±0.4 & 88.87±1.0 & 88.68±0.2 & 88.62±0.6 & 89.25±0.4  & 90.34±0.3 & 89.01±0.2 & 91.47±0.2 & 90.30±0.4 & \uline{92.12±0.3} & 78.39±0.2 & \uwave{92.10±0.2} & \textbf{93.97±0.2} \\
          &       & 60    & 90.78±0.1 & 89.20±0.8 & 90.25±0.1 & 89.19±0.9 & 89.46±0.6 & 90.64±0.3 & 88.85±0.2 & \uline{93.49±0.2} & 91.54±0.2 & 92.33±0.3 & 80.58±0.2 & \uwave{93.23±0.2} & \textbf{94.60±0.2} \\
\cmidrule{2-16}          & \multirow{3}[2]{*}{Mi-F1} & 20    & 89.94±0.5 & 90.16±0.9 & 89.67±0.1 & 88.72±2.6 & 90.78±0.3 & 91.97±0.2 & 89.71±0.3 & \uwave{93.06±0.4 } & 91.14±0.3 & 92.71±0.5 & 79.07±1.0 & \uline{93.18±0.4} & \textbf{94.68±0.2} \\
          &       & 40    & 90.37±0.4 & 89.47±0.9 & 89.14±0.2 & 89.22±0.5 & 89.92±0.4 & 91.97±0.2 & 89.61±0.3 & 91.77±0.2  & 90.79±0.4 & \uline{92.43±0.3} & 78.57±0.3 & \uwave{92.41±0.2} & \textbf{94.21±0.2} \\
          &       & 60    & 90.11±0.2 & 90.34±0.8 & 91.17±0.1 & 90.35±0.8 & 90.66±0.5 & 91.59±0.2 & 89.96±0.2 & \uline{94.13±0.2} & 92.35±0.2 & 93.05±0.3 & 81.33±0.1 & \uwave{93.95±0.2} & \textbf{95.13±0.2} \\
\cmidrule{2-16}          & \multirow{3}[2]{*}{AUC} & 20    & 98.05±0.4 & 98.07±0.6 & 97.69±0.0 & 96.99±1.4 & 97.75±0.3 & 98.32±0.1 & 97.77±1.2 & \uwave{99.09±0.1} & 98.74±0.1 & 98.90±0.1 & 93.25±0.1 & \uline{99.17±0.1} & \textbf{99.48±0.1} \\
          &       & 40    & 97.92±0.2 & 97.48±0.6 & 97.08±0.0 & 97.12±0.4 & 97.23±0.2 & 98.06±0.1 & 97.79±0.4 & \uwave{98.81±0.1} & 98.42±0.2 & 98.55±0.1 & 92.57±0.0 & \uline{98.95±0.0} & \textbf{99.12±0.1} \\
          &       & 60    & 98.24±0.2 & 97.96±0.5 & 98.00±0.0 & 97.76±0.5 & 97.72±0.4 & 98.59±0.1 & 98.26±0.3 & \uline{99.41±0.0} & 99.05±0.1 & 98.89±0.1 & 94.31±0.0 & \uwave{99.36±0.0} & \textbf{99.50±0.0} \\
    \midrule
    \multirow{9}[6]{*}{ACM} & \multirow{3}[2]{*}{Ma-F1} & 20    & 88.33±0.8 & 85.66±2.1 & 51.91±0.9 & 79.27±3.8 & 87.86±0.2 & 88.56±0.8 & 83.65±1.3 & \uwave{91.93±0.3} & 89.22±0.3 & 90.66±0.4 & 90.21±0.3 & \uline{92.84±0.2} & \textbf{93.38±0.1} \\
          &       & 40    & 87.45±0.2 & 87.47±1.1 & 62.41±0.6  & 80.23±3.3 & 86.23±0.8  & 87.61±0.5 & 85.86±0.6 & \uwave{91.35±0.3} & 89.04±0.4 & 90.15±0.6 & 90.83±0.2 & \uline{92.04±0.2} & \textbf{92.32±0.3} \\
          &       & 60    & 89.19±0.1 & 88.41±1.1 & 61.13±0.4 & 80.03±3.3 & 87.97±0.4 & 89.04±0.5 & 86.21±0.5 & \uwave{92.10±0.3} & 89.30±0.3 & 91.59±0.4 & 90.76±0.1 & \uline{92.11±0.3} & \textbf{92.54±0.1} \\
\cmidrule{2-16}          & \multirow{3}[2]{*}{Mi-F1} & 20    & 88.34±0.7 & 85.11±2.2 & 53.13±0.9 & 79.63±3.5 & 87.60±0.8 & 88.13±0.8 & 83.66±1.4 & \uwave{91.82±0.3} & 89.24±0.3 & 90.24±0.5 & 89.82±0.1 & \uline{92.61±0.2} & \textbf{93.27±0.1} \\
          &       & 40    & 87.24±0.2 & 87.21±1.2 & 64.43±0.6 & 80.41±3.0 & 86.02±0.9 & 87.45±0.5 & 86.01±0.7 & \uwave{91.33±0.3} & 89.08±0.4 & 90.18±0.6 & 90.66±0.2 & \uline{91.93±0.3} & \textbf{92.12±0.3} \\
          &       & 60    & 89.14±0.1 & 88.10±1.2 & 62.72±0.3 & 80.15±3.2 & 87.82±0.5 & 88.71±0.5 & 86.12±0.5 & \uline{91.99±0.3} & 88.97±0.5 & 91.34±0.4 & 90.66±0.1 & \uwave{91.88±0.3} & \textbf{92.35±0.1} \\
\cmidrule{2-16}          & \multirow{3}[2]{*}{AUC} & 20    & 96.76±0.2 & 93.47±1.5 & 71.66±0.7 & 91.47±2.3 & 96.72±0.3  & 96.49±0.3 & 94.07±0.2 & \uwave{98.43±0.2} & 97.82±0.1 & 97.69±0.1 & 98.01±0.2 & \uline{98.59±0.1} & \textbf{98.67±0.0} \\
          &       & 40    & 96.58±0.1 & 94.84±0.9 & 80.48±0.4 & 91.52±2.3  & 96.35±0.3 & 96.40±0.4 & 94.94±0.1 & \uwave{97.94±0.1} & 97.57±0.1 & 97.52±0.1 & 98.39±0.0 & \uline{98.29±0.3} & \textbf{98.57±0.1} \\
          &       & 60    & 97.05±0.0 & 94.68±1.4  & 79.33±0.4 & 91.41±1.9  & 96.79±0.2 & 96.55±0.3 & 95.34±0.2 & \uwave{98.40±0.2} & 97.40±0.2 & 97.87±0.1 & 98.05±0.0 & \uline{98.44±0.3} & \textbf{98.71±0.0} \\
    \midrule
    \multirow{9}[6]{*}{IMDB} & \multirow{3}[2]{*}{Ma-F1} & 20    & 35.70±2.7 & 27.50±1.5 & 40.18±0.7 & 28.62±2.9 & \uwave{50.99±1.5} & 36.38±0.2 & 16.99±0.4 & 49.19±0.3 & 40.12±1.0 & 46.55±0.4 & 42.28±0.1 & \uline{51.42±0.4} & \textbf{53.10±0.6} \\
          &       & 40    & 44.5±1.89 & 37.65±1.9 & 41.64±1.1 & 35.12±0.9 & \uline{53.26±1.5} & 44.66±0.3 & 17.52±0.3 & 49.78±0.5 & 41.31±0.8 & 45.86±1.0 & 52.90±0.2 & \uwave{53.11±0.4} & \textbf{57.07±0.4} \\
          &       & 60    & 48.3±1.5 & 46.87±1.6 & 45.54±0.9 & 35.99±1.0 & \uline{55.54±1.5} & 45.87±0.2 & 18.25±0.3 & 51.99±1.0 & 41.13±0.7 & 48.02±1.1 & 51.81±0.2 & \uwave{55.12±0.2} & \textbf{58.72±0.2} \\
\cmidrule{2-16}          & \multirow{3}[2]{*}{Mi-F1} & 20    & 21.72±6.1 & 37.08±0.6 & 41.22±0.8 & 38.09±1.0 & \uwave{51.56±1.5} & 42.13±0.1 & 34.21±0.3 & 50.86±0.3 & 42.73±0.9 & 48.39±0.6 & 46.04±0.2 & \uline{52.01±0.8} & \textbf{53.83±0.7} \\
          &       & 40    & 41.55±3.7 & 43.25±1.3 & 43.80±0.9 & 42.42±0.3 & \uwave{53.32±1.5} & 47.17±0.2 & 35.68±0.3 & 51.18±0.5 & 42.98±0.7 & 47.52±1.0 & 53.13±0.2 & \uline{53.65±0.4} & \textbf{57.62±0.4} \\
          &       & 60    & 44.62±2.5 & 49.80±1.0 & 47.68±0.7 & 44.78±0.4 & \uline{56.00±1.4} & 49.15±0.1 & 37.74±0.3 & 53.42±0.8 & 43.93±0.6 & 50.95±0.9 & 52.58±0.3 & \uwave{55.65±0.3} & \textbf{59.45±0.3} \\
\cmidrule{2-16}          & \multirow{3}[2]{*}{AUC} & 20    & 62.28±1.2 & 62.38±0.4 & 58.69±0.7 & 64.60±3.7 & 69.85±1.4 & 66.50±0.0 & 59.45±1.2 & 71.44±0.1 & 60.16±0.5 & 66.70±0.8 & \uwave{70.90±0.1} & \uline{72.96±0.1} & \textbf{75.64±0.1} \\
          &       & 40    & 63.02±0.6 & 66.67±0.5 & 62.41±0.6 & 65.44±0.2 & 70.51±1.5 & 66.75±0.0 & 59.76±2.7 & 70.24±0.3 & 61.95±0.5 & 67.22±0.6 & \uline{73.28±0.0} & \uwave{72.85±0.2} & \textbf{76.63±0.2} \\
          &       & 60    & 67.01±0.4 & 70.15±0.2 & 64.60±0.7 & 68.30±0.1 & \uwave{72.84±1.1} & 70.48±0.0 & 60.56±1.7 & 71.58±0.2 & 62.28±0.4 & 68.59±0.6 & 72.13±0.1 & \uline{74.46±0.1} & \textbf{77.93±0.2} \\
    \midrule
    \multirow{9}[6]{*}{Yelp} & \multirow{3}[2]{*}{Ma-F1} & 20    & 66.60±5.9 & 88.34±1.4 & 55.83±0.9 & 72.31±0.2 & 68.44±1.8 & 68.70±0.5 & 58.19±1.2 & 61.77±0.4 & \uwave{90.67±0.9} & 60.04±0.9 & 89.91±0.3 & \textbf{94.06±0.4} & \uline{93.29±0.3} \\
          &       & 40    & 68.66±3.7 & 87.82±0.3 & 57.91±1.2 & 75.89±0.1 & 71.87±1.5 & 68.50±1.4 & 58.82±0.3 & 64.88±0.3 & \uwave{90.89±1.4} & 61.75±1.4 & 90.74±0.2 & \textbf{94.19±0.1} & \uline{94.05±0.3} \\
          &       & 60    & 73.17±1.4 & 84.14±3.9 & 58.97±0.9 & 70.83±0.3 & 71.43±1.2 & 68.71±0.2 & 57.48±0.7 & 60.79±0.6 & \uwave{89.78±1.2} & 60.99±0.9 & 89.48±0.2 & \textbf{92.85±0.1} & \uline{92.60±0.1} \\
\cmidrule{2-16}          & \multirow{3}[2]{*}{Mi-F1} & 20    & 63.31±8.9 & 88.12±1.1 & 59.11±1.1 & 74.16±0.2 & 73.19±1.6 & 71.71±0.5 & 71.41±3.1 & 66.93±2.2 & \uwave{89.89±0.8} & 68.45±3.5 & 89.64±0.3 & \textbf{93.23±0.5} & \uline{92.54±0.3} \\
          &       & 40    & 67.13±5.7 & 87.57±0.3 & 62.15±1.4 & 77.27±0.1 & 74.37±1.9 & 71.43±1.6 & 70.57±5.2 & 68.16±1.5 & \uwave{90.25±1.1} & 66.82±4.3 & 90.08±0.2 & \textbf{93.33±0.2} & \uline{93.06±0.3} \\
          &       & 60    & 72.79±3.6 & 84.53±2.9 & 63.31±1.0 & 73.14±0.3 & 74.91±1.2 & 72.09±0.4 & 72.43±3.5 & 66.56±2.9 & \uwave{89.26±0.9} & 67.23±3.4 & 89.19±0.2 & \textbf{91.88±0.3} & \uline{91.53±0.3} \\
\cmidrule{2-16}          & \multirow{3}[2]{*}{AUC} & 20    & 82.17±4.3 & 96.51±0.8 & 76.64±0.7 & 88.67±0.1 & 84.47±5.1 & 89.40±2.6 & 81.51±4.2 & 84.63±0.7 & \uwave{97.40±0.3} & 80.34±4.1 & 97.27±0.1 & \textbf{98.54±0.1} & \uline{98.39±0.1} \\
          &       & 40    & 83.50±3.1 & 96.70±0.3 & 80.01±0.7 & 89.72±0.1 & 85.20±5.7 & 88.90±3.1 & 81.57±4.3 & 86.22±1.4 & \uwave{97.68±0.2} & 83.15±3.9 & 97.15±0.0 & \textbf{98.66±0.1} & \uline{98.58±0.1} \\
          &       & 60    & 86.81±1.7 & 95.38±0.5 & 80.94±0.6 & 88.04±0.1 & 88.22±2.1 & 87.06±2.7 & 81.35±2.7 & 83.52±2.7 & \uwave{97.12±0.3} & 83.47±3.3 & 96.26±0.2 & \textbf{98.28±0.1} & \uline{98.10±0.1} \\
    \bottomrule
    \end{tabular}%
    }
  \label{CLASSIFICATION}%
\end{table*}%

\textbf{Settings.} To ensure fair comparisons, we conduct 10 runs of all experiments and report the average results. For each dataset, we use the features of target nodes and set the final representation dimension to 64. For random-walk-based methods like Mp2vec, we set the number of walks per node to 40, the walk length to 100, and the window size to 5. For the homogeneous graph methods like GAT, DGI, GraphMAE, and GREET, we evaluate them on each meta-path subgraph and select the optimal results, following \cite{wang2021self,yu2023heterogeneous,tian2023heterogeneous}. A portion of the results for comparative methods is cited from \cite{tian2023heterogeneous}.

We initialize parameters using Kaiming initialization \cite{he2015delving} and train the model with Adam optimizer. We conduct experiments with different learning rates ranging from \{1e-3, 5e-4\} and penalty weights for L2-norm regularization from \{1e-3, 1e-4, 0\}. Early stopping is used with a patience of 10 epochs, stopping training if the total loss did not decrease for patience consecutive epochs. The temperature $\tau$ is adjusted from 0.3 to 1.0 with a step size of 0.1. We set the number of neighbors $K$ in the latent graphs from \{3, 5, 10\} and the number of positive samplers $k_{pos}$ from \{0, 1, 2, 3, 4\}. $k_{pos}$ with the value of 0 means that the set of positive samples consists solely of the node itself.
The sharpening parameter $\gamma$ is adjusted from \{1, 2\}. The filter order $r$ is set to 2 and $m$ is set to 1000.
The non-linear activation function used in this study is ELU$(\cdot)$. All experiments are implemented in the PyTorch platform using an AMD EPYC 7543 CPU and GeForce RTX 3090 24G GPU.

\subsection{Performance on Node Classification}

The node representations learned through unsupervised methods are used to train a linear classifier. To comprehensively assess our methods, we follow \cite{wang2021self} and select 20, 40, and 60 labeled nodes per class as the training set, while using 1000 nodes for validation and 1000 other nodes for testing in each dataset. We employ standard evaluation metrics, including Ma-F1, Mi-F1, and AUC, where higher values indicate better model performance. As shown in Table \ref{CLASSIFICATION}, among all the evaluated datasets, LatGRL achieves the best performance except for the Yelp dataset, where LatGRL-S outperforms it. Even though all other comparative methods employ full-graph training schemes, which often yield better results, LatGRL-S consistently ranks in the top three for each dataset. Additionally, the difference between LatGRL-S and LatGRL is relatively small, indicating the effectiveness of LatGRL-S. We can also observe that, in the majority of cases, excellent performance can be achieved when the number of training samples per class is only 20. This finding substantiates the high quality of the learned representations.

In particular, for Yelp and IMDB datasets, which exhibit low overall semantic homophily ratios, existing UHGRL methods such as HeCo, MEOW, and HGMAE perform poorly, while LatGRL outperforms them significantly. This emphasizes the challenge posed by semantic heterophily and the effectiveness of LatGRL. Moreover, GREET, despite accounting for edge heterophily, demonstrates inferior performance attributed to its single-view limitation, which highlights the necessity of leveraging multi-view information from diverse meta-paths in heterogeneous graphs.

\begin{table}[]
  \centering
  \caption{Quantitative results ($\%$) on node clustering.}
  \resizebox{\linewidth}{!}
  {
    \begin{tabular}{c|cc|cc|cc|cc}
    \toprule
    Datasets & \multicolumn{2}{c|}{DBLP} & \multicolumn{2}{c|}{ACM} & \multicolumn{2}{c|}{IMDB} & \multicolumn{2}{c}{Yelp} \\
    \midrule
    Metrics & NMI   & ARI   & NMI   & ARI   & NMI   & ARI   & NMI   & ARI \\
    \midrule
    GAT (2018) & 71.32 & 75.78 & 59.59 & 62.49 & 5.74  & 4.81  & 44.14 & 45.88 \\
    HAN (2019) & 67.98 & 74.02 & 60.63 & 64.28 & \uwave{9.53} & 9.08  & 64.21 & 67.55 \\
    \midrule
    Mp2vec (2017) & 73.55 & 77.70  & 48.43 & 34.65 & 5.87  & 4.37  & 38.90  & 42.49 \\
    DGI (2019) & 59.23 & 61.85 & 51.73 & 41.16 & 6.97  & 8.12  & 39.03 & 42.53 \\
    DMGI (2020) & 70.06 & 75.46 & 51.66 & 46.64 & 9.41   & \uwave{10.12} & 36.95 & 32.56 \\
    HeCo (2021) & 74.51 & 80.17 & 56.87 & 56.94 & 2.14  & 2.79  & 39.02 & 42.53 \\
    GraphMAE (2022) & 67.79 & 71.01 & 53.93 & 53.09 & 4.12  & 2.51  & 39.08 & 43.57 \\
    MEOW (2023) & 75.46 & 81.19 & 66.21 & 71.17 & 8.23  & 8.97  & 41.88 & 40.72 \\
    DuaLGR (2023) & 73.23 & 79.64 & 61.36 & 65.07 & 2.91  & 2.29  & \uwave{69.31} & \uwave{71.69} \\
    HGMAE (2023) & \uwave{76.92} & \uwave{82.34} & \uwave{66.68} & \uwave{71.51} & 5.55  & 5.86  & 38.95 & 42.6 \\
    \midrule
    GREET (2023) & 47.06 & 49.36 & 65.56 & 69.32 & 7.66  & 7.85  & 42.53 & 43.76 \\
    \midrule
    \textbf{LatGRL-S} & \uline{78.37} & \uline{84.04} & \uline{71.71} & \uline{74.13} & \uline{9.92} & \uline{10.88} & \uline{73.92} & \uline{77.72} \\
    \midrule
    \textbf{LatGRL} & \textbf{81.35} & \textbf{86.07} & \textbf{72.52} & \textbf{76.75} & \textbf{10.64} & \textbf{12.13} & \textbf{74.24} & \textbf{77.81} \\
    \bottomrule
    \end{tabular}%
   }
  \label{CLUSTERING}%
\end{table}%

\begin{figure*}[!t]
    \centering
    \subfloat[Mp2vec:-0.0059]{\includegraphics[width=0.17\linewidth]{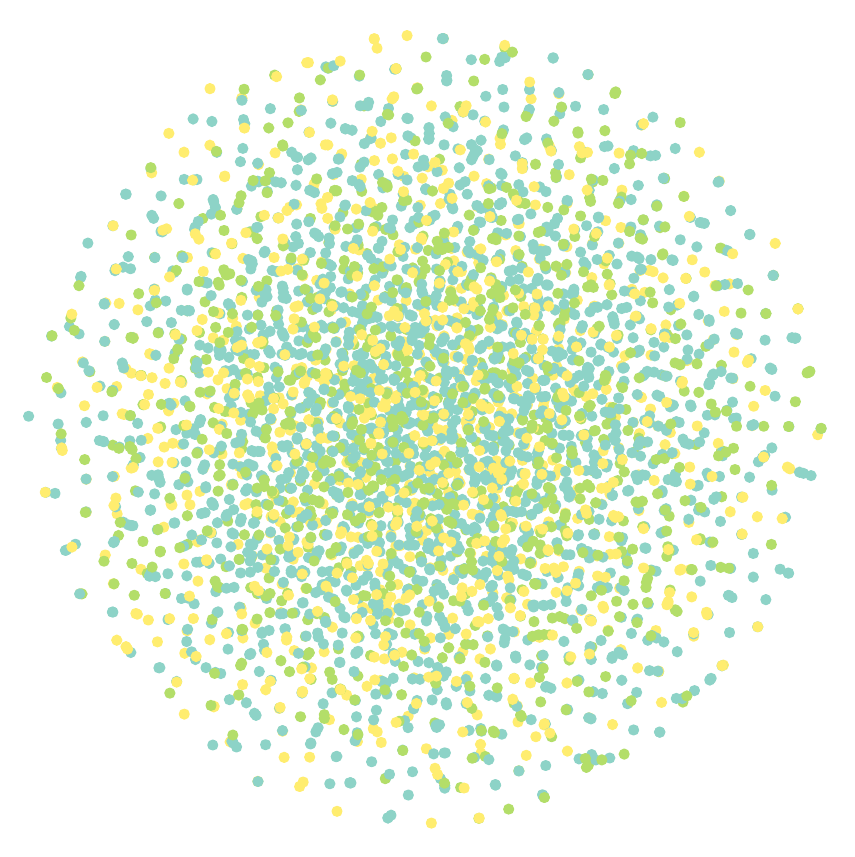}\label{Vis-mp2vec}}
    \hfill
    \subfloat[DMGI:0.3129]{\includegraphics[width=0.17\linewidth]{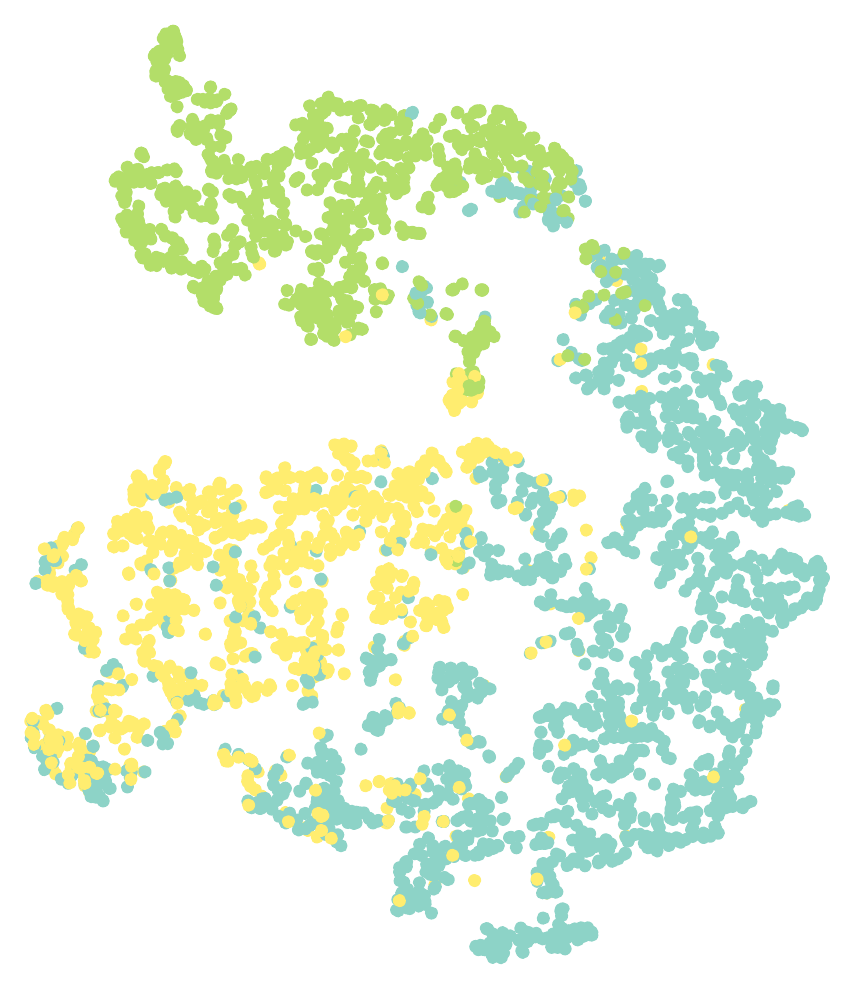}\label{Vis-dmgi}}
    \hfill
    \subfloat[HeCo:0.3473]{\includegraphics[width=0.17\linewidth]{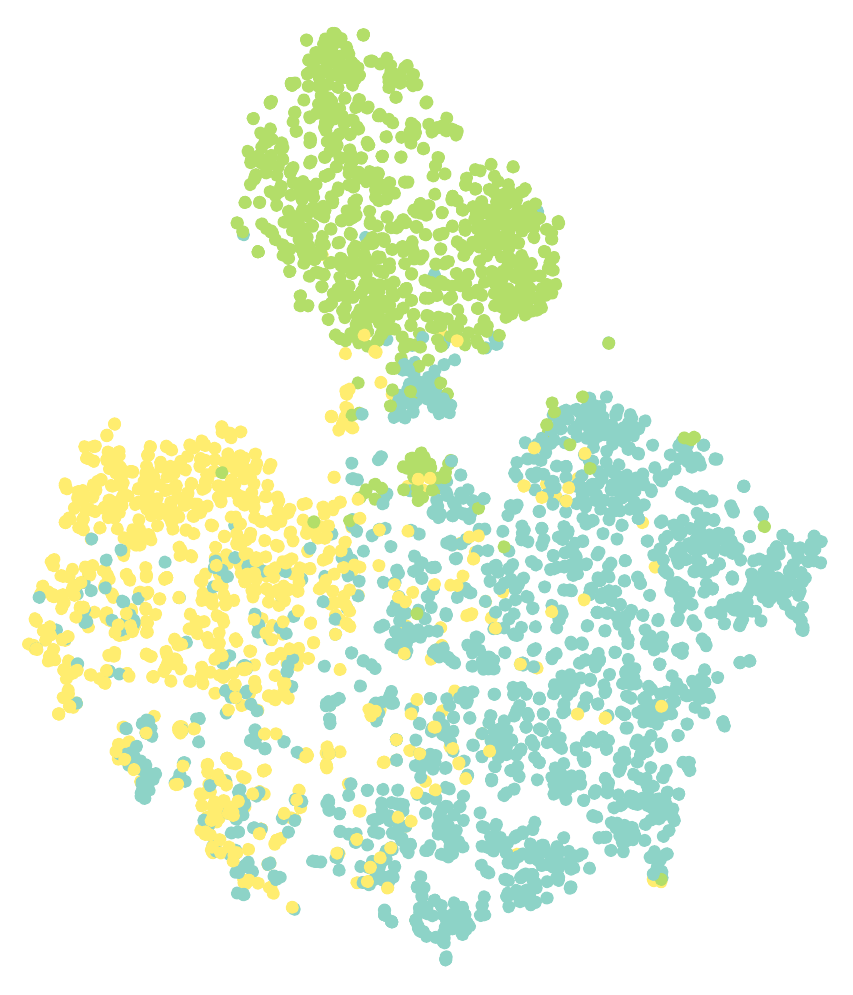}\label{Vis-heco}}
    \hfill
    \subfloat[HGMAE:0.3436]{\includegraphics[width=0.17\linewidth]{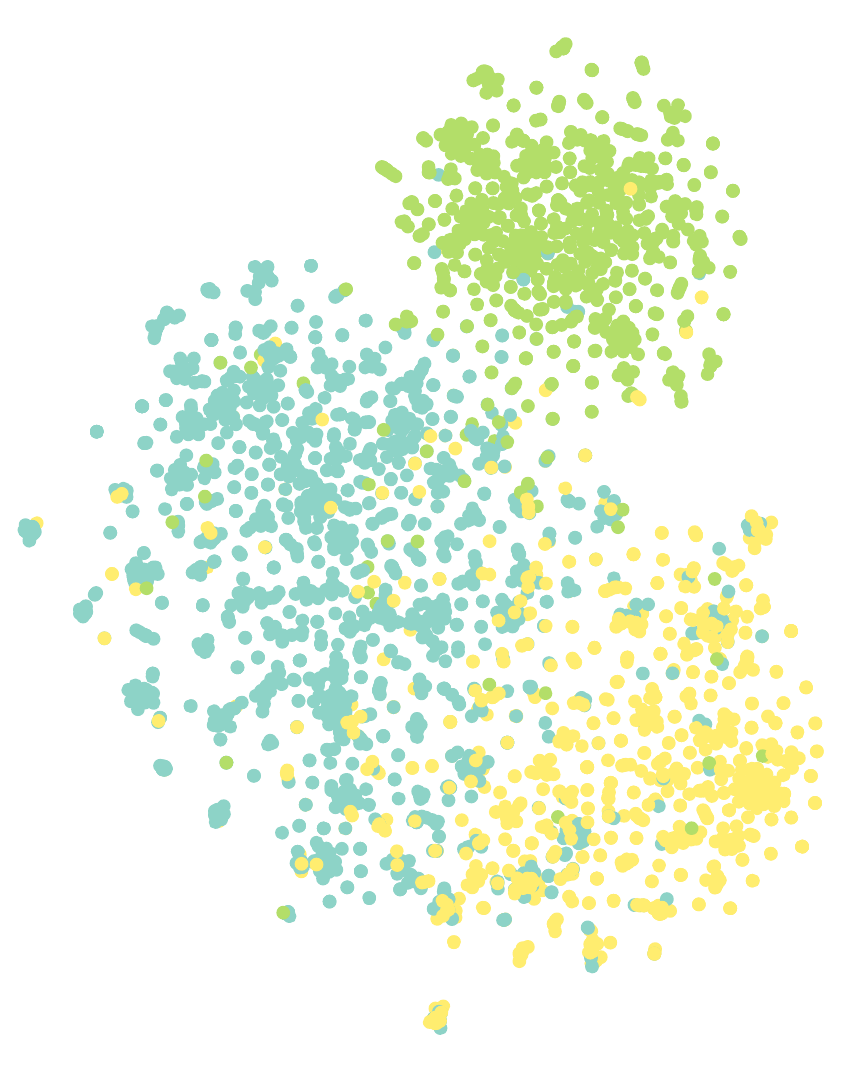}\label{Vis-hgmae}}
    \hfill
    \subfloat[LatGRL:0.4170]{\includegraphics[width=0.17\linewidth]{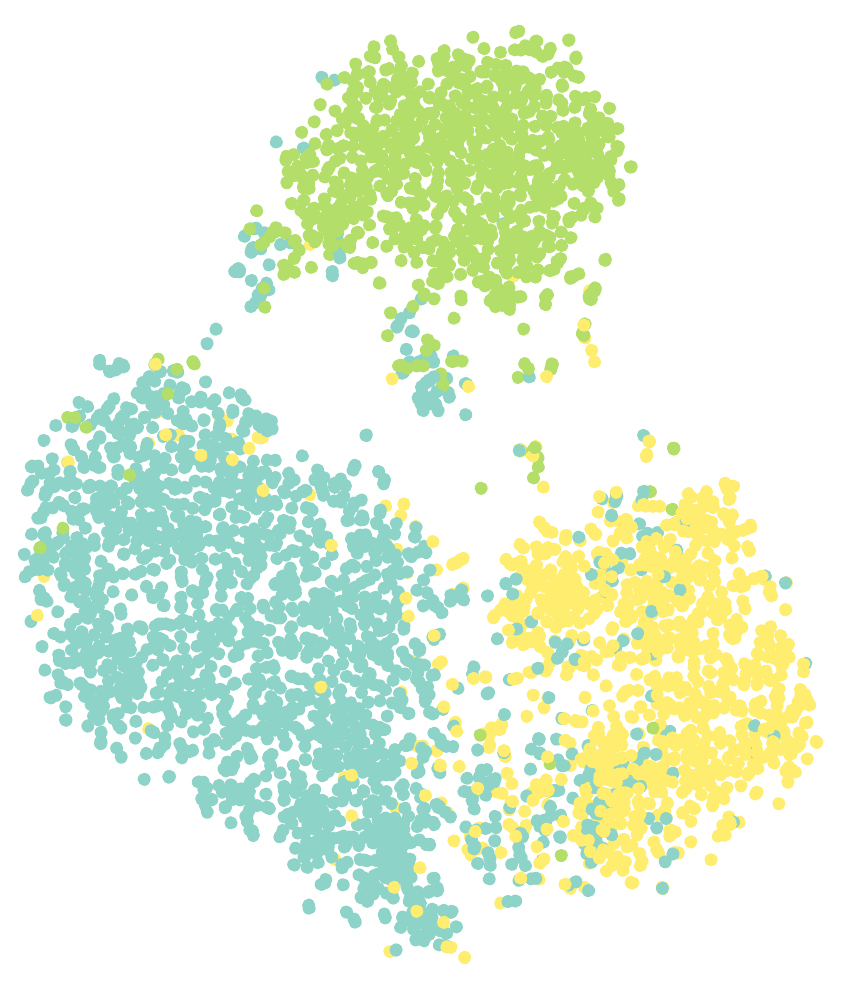}\label{Vis-LatGRL}}
    \caption{Visualization of the learned node representation on ACM. The corresponding Silhouette scores are also given.}
    \label{Visual}
\end{figure*}

\subsection{Performance on Node Clustering}

In this task, we further perform the k-means algorithm to the learned representations of all nodes and adopt normalized mutual information (NMI) and adjusted rand index (ARI) to assess the quality of the clustering results. For both metrics, the larger, the better. To alleviate the instability due to different initial values, we repeat the process 10 times and report average results in Table \ref{CLUSTERING}.
We can observe that LatGRL outperforms existing methods in all datasets, with LatGRL-S consistently securing the second-highest ranking. Remarkably, when it comes to the Yelp dataset, most existing UHGRL methods fail to achieve exceptional discriminative category performance. However, LatGRL stands out by surpassing them by a considerable margin, which shows the strong capability of LatGRL in acquiring category discriminative information.

To provide an intuitive evaluation, we visualize learned node representations on the ACM dataset through t-SNE in Fig. \ref{Visual}. Different colors mean different labels. We could observe that Mp2vec completely mixes the representations of different categories, indicating its lack of effective category discrimination capability. For DMGI, the clusters lack tightness. As for HeCo and HGMAE, although some categories are correctly classified, they still have many overlapped data points and blurred boundaries. LatGRL correctly separates nodes of different categories and exhibits clear boundaries. Furthermore, we compute the Silhouette score for the formed clusters, where a larger value indicates better clustering performance. LatGRL surpasses all other methods, demonstrating its effectiveness and the superior category discrimination capability of its learned representations.

\subsection{Ablation Study}

\textit{1) Effectiveness of Each Loss:} To examine the effectiveness of each component of LatGRL, we conduct experiments on variants of LatGRL. We first remove key components from latent graphs guided learning to examine their effectiveness, i.e., homophilic latent graph guided learning (w/o Hom), heterophilic latent graph guided learning (w/o Het), and original features maintenance (w/o OFM). We report the results of 40 labeled nodes per class in Table \ref{ABLATION}. The results show that both components are critical to LatGRL, and the guidance of latent graphs seems to contribute more. 

\textit{2) Effectiveness of Each Module:} Note that LatGRL employs adaptive dual-frequency semantic fusion. So we remove low-frequency information (w/o LF) and high-frequency information (w/o HF), which means that only one frequency of semantic information is used for fusion. We also replace the adaptive fusion mechanism with fixed semantic fusion used in previous UHGRL methods (w/o AF), where all nodes share the same fusion coefficients. It is obvious that low-frequency information is more crucial for most datasets. However, the Yelp dataset exhibits poorer performance when high-frequency information is missing. This discrepancy is consistent with its lower overall semantic homophily ratio. Moreover, the adaptive fusion mechanism, compared to the sharing of coefficients among all nodes, actually generates better representations.

\begin{table}[t]
  \centering
  \caption{Performance ($\%$) of LatGRL and its variants.}
    \resizebox{\linewidth}{!}
  {
    \begin{tabular}{c|l|c|c|c|c}
    \toprule
    \multicolumn{1}{l|}{Metrics} & Variants & DBLP  & ACM   & IMDB  & Yelp \\
    \midrule
    \multirow{7}[6]{*}{Ma-F1} & \multicolumn{1}{c|}{\textbf{LatGRL}} & \textbf{93.97±0.2} & \textbf{92.32±0.3} & \textbf{57.07±0.4} & \textbf{94.05±0.3} \\
\cmidrule{2-6}          & w/o Hom & 93.86±0.2 & 85.69±0.5 & 54.52±0.5 & 93.19±0.2 \\
          & w/o Het & 93.46±0.4 & 90.26±0.4 & 55.14±0.4 & 93.72±0.2 \\
          & w/o OFM & 93.20±0.2 & 91.11±0.3 & 55.93±0.5 & 93.74±0.3 \\
\cmidrule{2-6}          & w/o LF & 44.08±1.0 & 47.86±1.1 & 38.29±0.4 & 86.75±0.4 \\
          & w/o HF & 92.98±0.2 & 88.69±0.5 & 56.14±0.6 & 65.46±2.5 \\
          & w/o AF & 93.82±0.2 & 91.70±0.3 & 54.56±0.5 & 93.85±0.3 \\
    \midrule
    \multirow{7}[6]{*}{Mi-F1} & \multicolumn{1}{c|}{\textbf{LatGRL}} & \textbf{94.23±0.2} & \textbf{92.12±0.3} & \textbf{57.62±0.4} & \textbf{93.06±0.3} \\
\cmidrule{2-6}          & w/o Hom & 94.13±0.2 & 85.75±0.5 & 55.18±0.6 & 92.35±0.2 \\
          & w/o Het & 93.62±0.4 & 90.15±0.4 & 55.68±0.6 & 92.82±0.3 \\
          & w/o OFM & 93.58±0.2 & 90.80±0.4 & 56.58±0.6 & 92.88±0.3 \\
\cmidrule{2-6}          & w/o LF & 43.58±0.9 & 48.94±1.0 & 38.76±0.6 & 84.15±0.4 \\
          & w/o HF & 93.29±0.2 & 88.29±0.5 & 56.94±0.8 & 71.61±3.5 \\
          & w/o AF & 94.05±0.2 & 91.50±0.3 & 55.57±0.5 & 92.81±0.4 \\
    \bottomrule
    \end{tabular}%
    }
  \label{ABLATION}%
\end{table}%

\begin{table}[htbp]
   \centering
  \caption{Similarity search ($\%$) on low NHR nodes of ACM.}
    \resizebox{\linewidth}{!}
  {
    \begin{tabular}{c|cccc|cccc}
    \toprule
    Method & Raw   & GCN   & HAN   & LatGRL & w/o Hom & w/o Het   & w/o LF    & w/o HF \\
    \midrule
    Sim@5 & 68.9  & 77.1  & 80.5  & \textbf{86.5}  & 82.2  & 85.5  & 43.4  & 84.8 \\
    Sim@10 & 67.5  & 76.8  & 80.4  & \textbf{86.9}  & 81.3  & 85.2  & 41.7  & 83.9 \\
    \bottomrule
    \end{tabular}%
    }
  \label{low_NHR}%
\end{table}%

\begin{table}[htbp]
   \centering
  \caption{Performance ($\%$) with different numbers of meta-paths.}
    \resizebox{\linewidth}{!}
  {
    \begin{tabular}{cccccc}
    \toprule
    \multirow{2}[4]{*}{Datasets} & \multirow{2}[4]{*}{Meta-path} & \multicolumn{2}{c}{Node Classification} & \multicolumn{2}{c}{Node Clustering} \\
\cmidrule{3-6}          &       & Ma-F1 & Mi-F1 & NMI   & ARI \\
    \midrule
    \multirow{3}[2]{*}{ACM} & PAP   & 89.77±0.2 & 89.79±0.2 & 60.58 & 65.31 \\
          & PSP   & 86.63±0.4 & 86.19±0.5 & 48.89 & 48.92 \\
          & PAP \& PSP & \textbf{93.38±0.1} & \textbf{93.27±0.1} & \textbf{72.52} & \textbf{76.75} \\
    \midrule
    \multirow{4}[2]{*}{Yelp} & BUB \& BSB & 92.56±0.2 & 92.02±0.2 & 71.48 & 73.49 \\
          & BUB \& BLB & 90.86±0.4 & 89.75±0.4 & 65.98 & 68.16 \\
          & BSB \& BLB & 92.77±0.3 & 91.81±0.3 & 71.43 & 74.57 \\
          & BUB \& BSB \& BLB & \textbf{93.29±0.3} & \textbf{92.54±0.3} & \textbf{74.24} & \textbf{77.81} \\
    \bottomrule
    \end{tabular}%
    }
  \label{num_metapath}%
\end{table}%

\textit{3) Effectiveness on Low NHR Nodes:} To further demonstrate the success of LatGRL on nodes with low NHR, we compare the performance of LatGRL and baselines in the similarity search task, as shown in Table \ref{low_NHR}. We select 200 nodes with the lowest NHR under the PAP meta-path of the ACM dataset. This task involves calculating the cosine similarity between the learned representations of each low NHR node and all other nodes in the entire graph. For each low NHR node, we select a certain number of most similar nodes and calculate the proportion of nodes with the same label. This task effectively measures the category discrimination ability of the learned node representations. ``Raw" denotes the original node features. It can be observed that despite the lack of label information, the representations learned by LatGRL still have superior category discrimination capability, even surpassing supervised methods. The right half of Table \ref{low_NHR} shows the performance of different variants of LatGRL on low NHR nodes. Each variant represents the removal of a key component. We can observe that removing low-pass graph filtering causes the most significant performance degradation, highlighting the importance of low-frequency information. However, considering that most existing GNN encoders can also be equivalent to low-pass filters \cite{bo2021beyond}, our proposed latent graphs guided learning becomes more pivotal. Especially, the homophilic latent graph plays a significant role in improving the performance of low NHR nodes.

\textit{4) Effectiveness of Each Meta-path:} Table \ref{num_metapath} shows the impact of the number of meta-paths on node classification and clustering. Firstly, different meta-paths have varying importance. For instance, in the ACM dataset, PAP is more significant. Secondly, the overall performance improves with more meta-paths. The performance with complete meta-paths consistently surpasses other variants, indicating that each meta-path provides indispensable complementary information.

\subsection{Performance on Large-scale Dataset}

To evaluate the scalability of LatGRL-S, we conduct experiments on a large heterogeneous graph dataset, ogbn-mag. In particular, the ogbn-mag dataset consists of 349 categories, making it challenging for unsupervised learning. The dimension of node representation is set to 512. The number of anchor nodes ($m$) is set to 5000 and the batch size is set to 5120. Regarding comparative methods, MLP and GAT are selected as supervised baselines. Furthermore, we consider four unsupervised graph learning methods: DGI, GCA\cite{zhu2021graph}, GIC\cite{mavromatis2021graph}, and SUGRL\cite{mo2022simple}. 
% The data partitioning follows the official training (85.5\%), validation (8.8\%), and testing (5.7\%) sets in this study. 
The experimental results of these comparative methods are copied from \cite{mo2022simple}.

\begin{figure}[t]
		\centering
		\includegraphics[width=8cm]{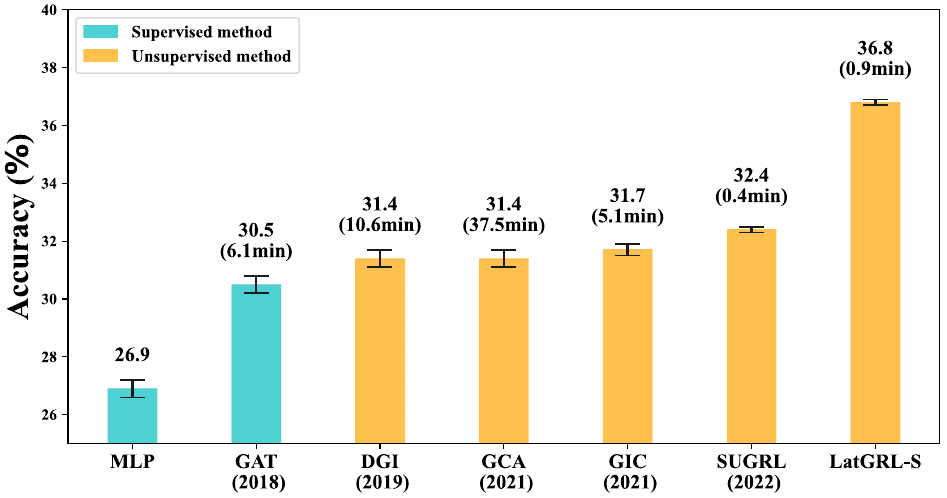}
	\caption{The experimental results on ogbn-mag.}
	\label{mag}
\end{figure}

Fig. \ref{mag} shows the accuracy and representation training time. Most existing unsupervised graph learning methods have demonstrated limited performance on it. However, LatGRL-S still outperforms all comparative methods by a significant margin, and its training time for representation is much lower than that for all of them except SUGRL, which demonstrates the effectiveness and scalability of LatGRL-S.

\begin{figure*}[!t]
    \centering
    \subfloat[DBLP]{\includegraphics[width=0.24\linewidth]{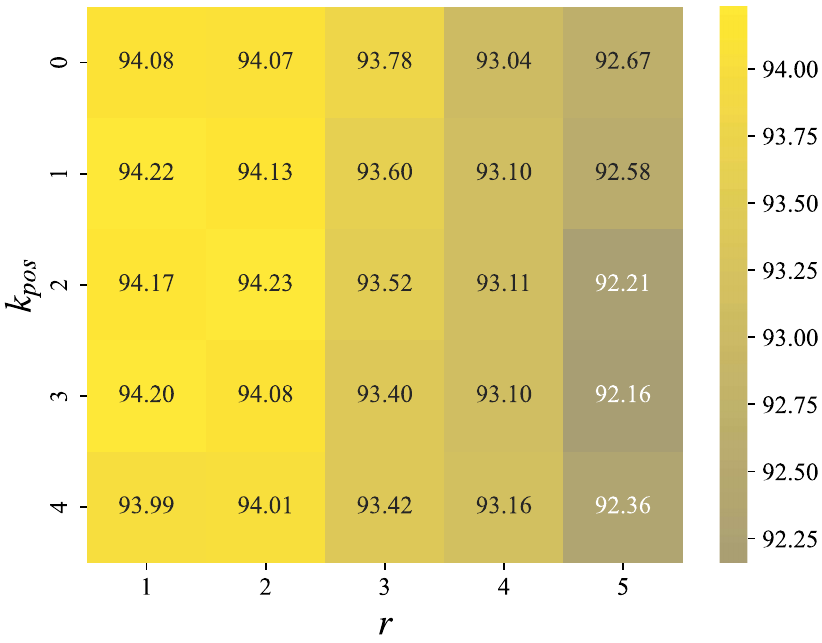}\label{DBLP_40}}
    \hfill
    \subfloat[ACM]{\includegraphics[width=0.24\linewidth]{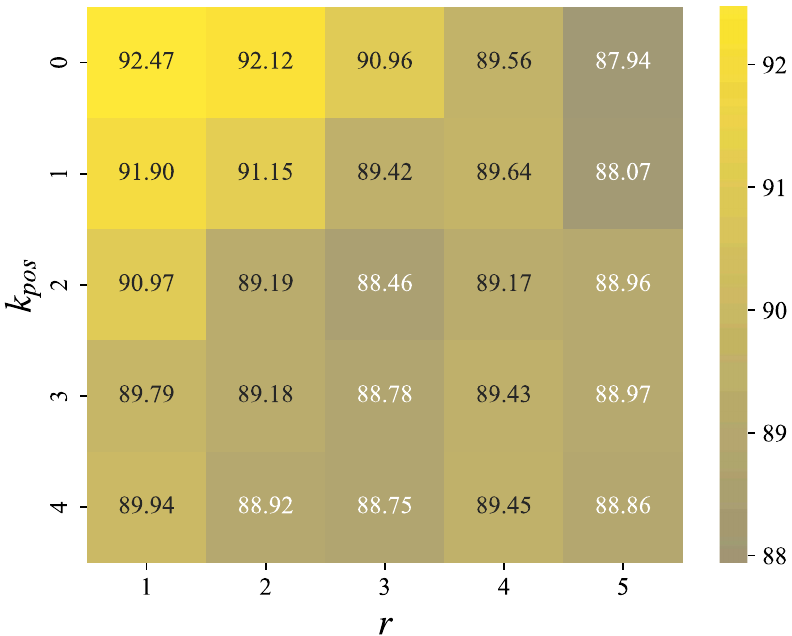}\label{ACM_40}}
    \hfill
    \subfloat[IMDB]{\includegraphics[width=0.24\linewidth]{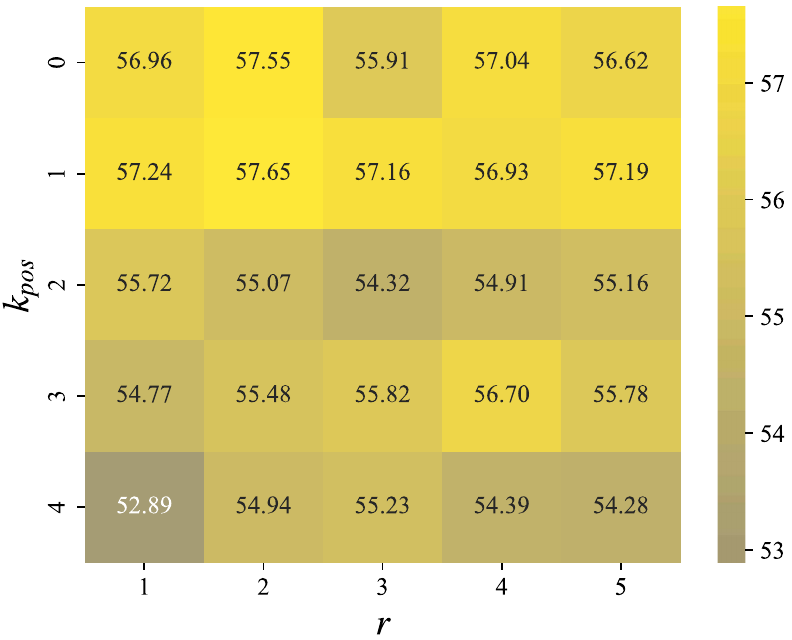}\label{IMDB_40}}
    \hfill
    \subfloat[Yelp]{\includegraphics[width=0.24\linewidth]{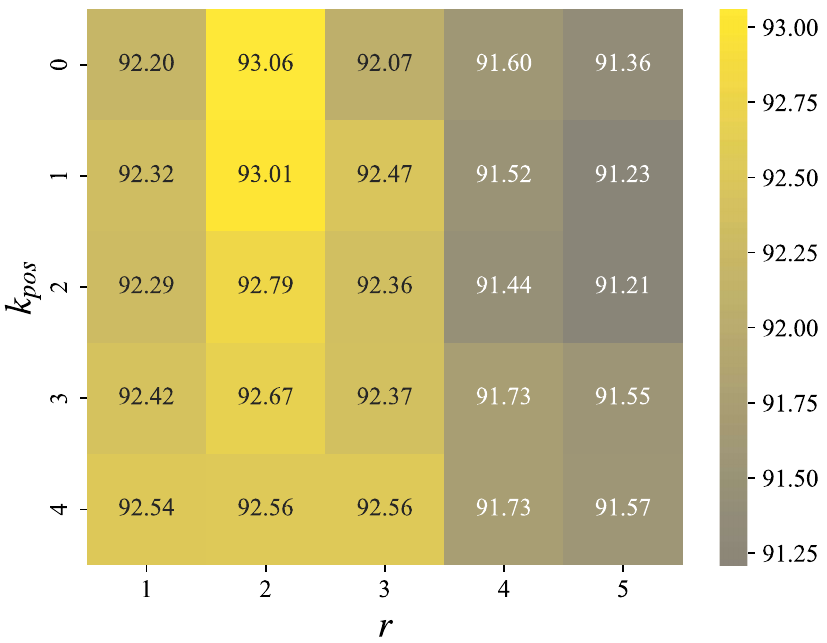}\label{YELP_40}}
    \caption{The influence of $r$ and $k_{pos}$ (metric: Mi-F1 (\%)).}
    \label{40-MI}
\end{figure*}

% \begin{figure*}[t]
%     \centering
%     \subfloat[DBLP]{\includegraphics[width=0.24\linewidth]{fig/dblp_K.pdf}\label{DBLP_K}}
%     \hfill
%     \subfloat[Yelp]{\includegraphics[width=0.24\linewidth]{fig/yelp-K.pdf}\label{YELP_K}}
%     \caption{The influence of $K$ (metric: Mi-F1 (\%)).}
%     \label{K_MI}
% \end{figure*}

\subsection{Hyper-parameter Analysis}

We further analyze the impact of two important hyper-parameters in LatGRL: the filtering order $r$ and the number of positive samples $k_{pos}$. Fig. \ref{40-MI} illustrates the results on the Micro-F1 scores with 40 labeled nodes per class. It can be observed that when the filtering order is relatively small, better results can be achieved. As the filtering order increases, the performance tends to decrease on most datasets. Smaller filtering orders also improve computational efficiency in practical applications. 

Furthermore, in the ACM and IMDB datasets, as $k_{pos}$ increases, the decrease in model performance becomes more pronounced. This could be attributed to the fact that higher values of $k_{pos}$ result in a decrease in the precision of the positive samples extracted. In contrast, the model is not sensitive to changes in $k_{pos}$ in the remaining two datasets.

% Fig. \ref{K_MI} reflects the influence of the neighbor $K$ of latent graphs. As the value of $K$ ranges from 3 to 20, the performance exhibits a consistent level of stability, with a variation range of no more than $1\%$, which suggests that the performance of LatGRL is not strongly influenced by changes of $K$.

\begin{figure*}[!t]
    \centering
    \subfloat[ACM]{\includegraphics[width=0.3\linewidth]{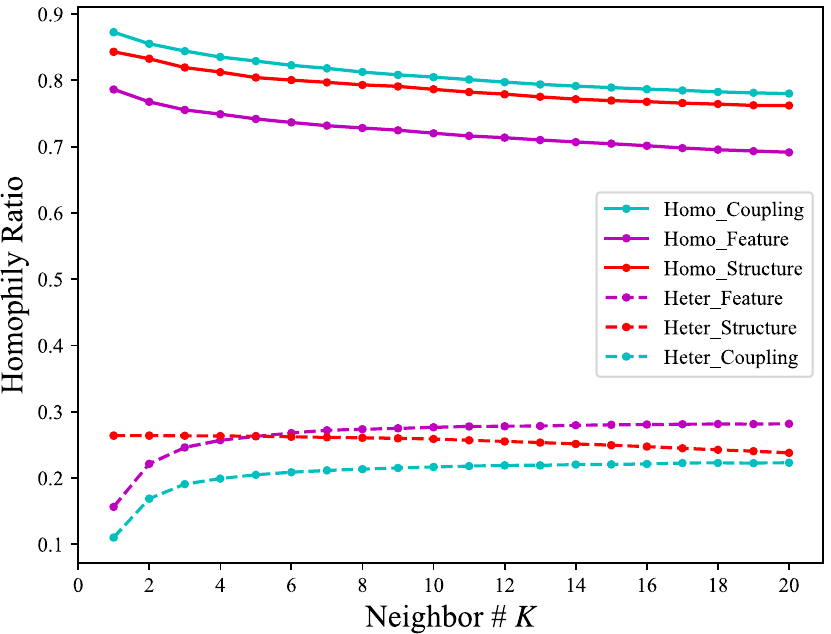}\label{ACM-LG}}
    \hfill
    \subfloat[Yelp]{\includegraphics[width=0.3\linewidth]{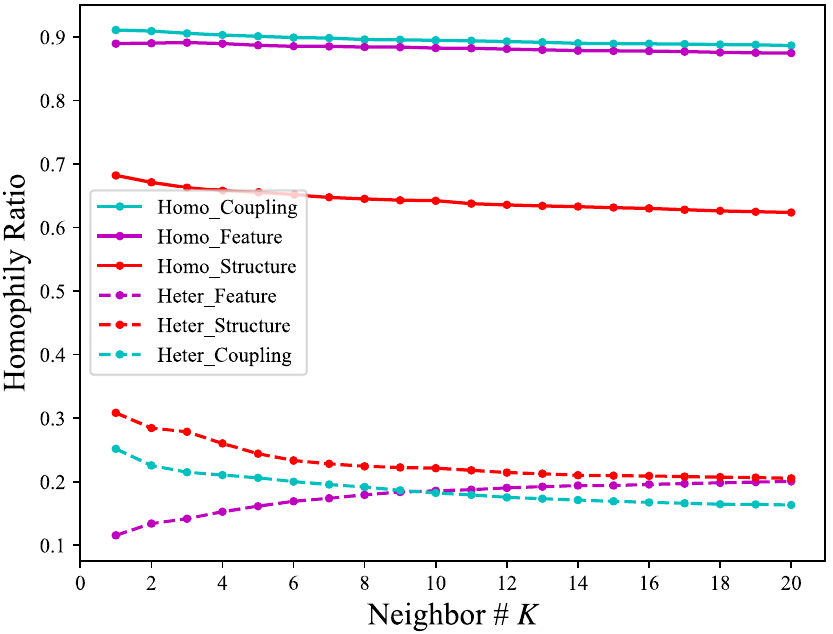}\label{YELP-LG}}
    \hfill
    \subfloat[DBLP]{\includegraphics[width=0.3\linewidth]{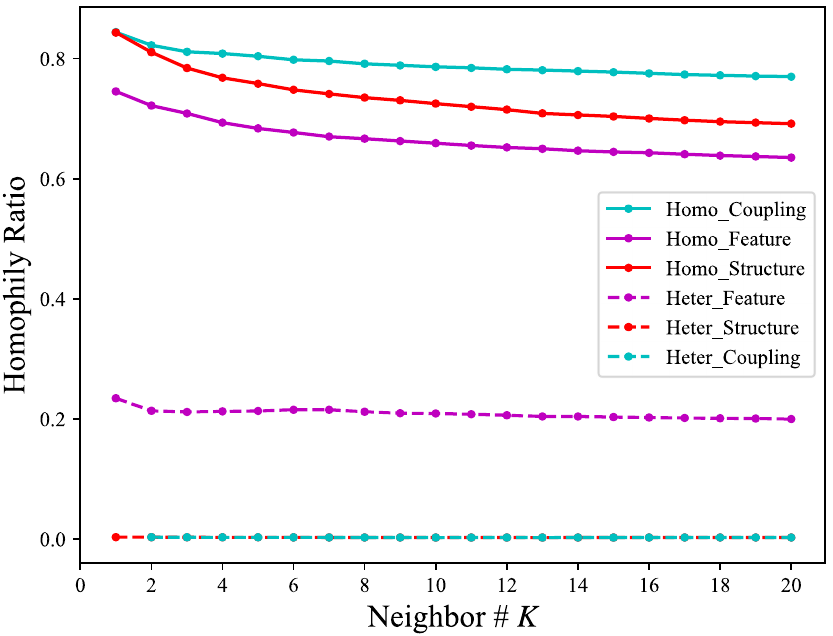}\label{DBLP-LG}}
    \caption{The superiority of our proposed similarity measurement that couples structures and features.}
    \label{LG-HR}
\end{figure*}

\subsection{Latent Graph Homophily Analysis}\label{6-G}

We conduct a comprehensive analysis to showcase the HR of latent graphs. Table \ref{HR_LATENT_GRAPHS} presents the HR of the latent graphs constructed in our study. For ACM and Yelp datasets, the HR of the homophilic latent graph surpasses the MHR based on any meta-path. Moreover, the HR of the homophilic latent graph in DBLP is very close to the highest MHR observed among meta-paths. Conversely, the HR of the heterophilic latent graph is consistently lower than the MHR of any meta-path in any dataset. In DBLP, it even approaches zero. Concerning the IMDB dataset, the general low semantic homophily ratios and the weak discriminative capability of the original features shown in Table \ref{HR-TABLE} may have jointly affected the HR of the latent graphs. However, LatGRL still outperforms existing UHGRL methods by a large margin in classification results.

Furthermore, we investigate the superiority of the approach that integrates global structures and features in capturing categorical information. Fig. \ref{LG-HR} presents the HR curves of the latent homophilic and heterophilic graphs constructed using our coupled mining method, denoted as $Homo\_Coupling$ and $Heter\_Coupling$ compared to the HR curves of graphs constructed solely based on structure or feature similarity, denoted as $Homo\_Structure$, $Heter\_Structure$ and $Homo\_feature$, $Heter\_feature$, respectively.
The curves illustrate the variation of HR with the number of neighbors $K$. In the HR curves for the homophilic latent graph, our coupled mining method outperforms using only the structure or features approach for any given dataset. Additionally, as $K$ increases, the homophilic latent graph with the coupled mining method consistently maintains a higher HR level, especially evident in the DBLP dataset. On the other hand, the HR for the heterophilic latent graph with coupled mining method consistently remains very low. These findings indicate the superiority of our coupled mining method in capturing categorical information and its insensitivity to changes in $K$.

\begin{table}[t]
\centering
\caption{The HR ($\%$) of the latent graphs (LG) compared to the MHR ($\%$) of the original heterogeneous graphs.}
% \resizebox{\linewidth}{!}
% {
\scalebox{0.75}{
\renewcommand{\arraystretch}{1.25}
\begin{tabular}{|c|c|c|c|}
\hline
Datasets & \begin{tabular}[c]{@{}c@{}}  MHR  \end{tabular}    & \begin{tabular}[c]{@{}c@{}}HR of\\ Homophilic LG\end{tabular} & \begin{tabular}[c]{@{}c@{}}HR of\\ Heterophilic LG\end{tabular} \\ \hline
ACM      & \begin{tabular}[c]{@{}c@{}}PAP: 80.85\\ PSP: 63.93\end{tabular}                  & 84.41                                                                   & 19.08                                                                     \\ \hline
IMDB     & \begin{tabular}[c]{@{}c@{}}MDM: 61.41\\ MAM: 44.43\end{tabular}                  & 47.85                                                                   & 26.49                                                                     \\ \hline
DBLP     & \begin{tabular}[c]{@{}c@{}}APA: 79.88\\ APCPA: 66.97\\ APTPA: 32.45\end{tabular} & 78.62                                                                   & 0.25                                                                      \\ \hline
Yelp     & \begin{tabular}[c]{@{}c@{}}BSB: 64.08\\ BUB: 44.97\\ BLB: 38.76\end{tabular}     & 89.47                                                                   & 18.23                                                                     \\ \hline
\end{tabular}
}
% }
  \label{HR_LATENT_GRAPHS}%
\end{table}

\section{Related Work}

\subsection{Unsupervised Heterogeneous Graph Learning}
UHGRL aims to learn low-dimensional node representations on heterogeneous graphs without labeled data. Early methods utilize structure information to maximize the representation similarity of proximal nodes within the same link\cite{chen2018pme}, the same random walk\cite{dong2017metapath2vec}, or the same subgraph\cite{zhang2020mg2vec}. With the development of deep representation learning, contrastive learning has been incorporated into heterogeneous graphs, achieving unsupervised representation learning by maximizing mutual information between contrastive views \cite{le2020contrastive}. Recently, DMGI\cite{park2020unsupervised} extends the mutual information maximization of local and global representations in DGI\cite{velivckovic2018deep} and adds semantic consistency constraints. Heco\cite{wang2021self}, MEOW\cite{yu2023heterogeneous}, HGCML\cite{hgcml}, and BMGC\cite{shen2024balanced} utilize multi-view information in heterogeneous graphs to facilitate contrastive learning. DuaLGR\cite{ling2023dual} benefits from the reconstruction of both structures and features in multi-view graph clustering. BTGF \cite{qian2024upper} develops a novel filter to handle multi-relational graphs. HGMAE\cite{tian2023heterogeneous} introduces the masked autoencoder into heterogeneous graphs first. However, due to the low-pass filtering of the encoding mechanisms and the structure strong dependence of the contrastive views they used, existing UHGRL methods inevitably cause representation smoothing of neighboring nodes with different categories, which is extremely detrimental for solving semantic heterophily in heterogeneous graphs.

\subsection{Heterophily and Graph Neural Network}
Recently, many works have discussed the performance degradation of GNNs on non-homophilic graphs and proposed some solutions. Geom-GCN\cite{pei2019geom} expands the node neighborhood by mining the geometric relationships of the input graph. FAGCN\cite{bo2021beyond}, SPCNet \cite{li2024simplified}, and ACM\cite{luan2022revisiting} adopt adaptive message passing through graph filtering on the original graph structure. In addition, \cite{yan2022two} has discussed the relationship between heterophily and over-smoothing in GNNs. It should be noted that, in unsupervised graph learning, heterophily has also received attention in recent studies like MUSE\cite{yuan2023muse}, RGSL\cite{xie2024robust}, and GREET\cite{liu2023beyond}. However, most studies have focused mainly on homogeneous graphs. Although there exists some research that has explored homophily/heterophily in heterogeneous graphs \cite{guo2023homophily,li2023hetero}, it focuses only on the meta-path-level, disregarding the diversity of heterophily at the node level. Moreover, their solutions are heavily dependent on node labels. On the contrary, we are the first to tackle the issue of heterophily in unsupervised heterogeneous graph learning, which is more challenging.

\section{Conclusion}

In this paper, we delve into an extensive analysis of semantic heterophily in heterogeneous graphs and present a pioneering effort to resolve it in an unsupervised manner. We propose the concept of semantic homophily and heterophily, along with the corresponding evaluation metrics, and conduct an empirical study to analyze the prevalent manifestation of semantic heterophily. To overcome these challenges, we propose a novel unsupervised heterogeneous graph representation learning framework called LatGRL, which utilizes a similarity mining approach that couples global structures and features to construct homophilic and heterophilic latent graphs to guide representation learning. LatGRL also incorporates an adaptive dual-frequency semantic fusion mechanism with dual-pass graph filtering to handle semantic heterophily. Extensive experiments on four public datasets and a large-scale dataset demonstrate that it outperforms existing state-of-the-art models, confirming its effectiveness and high efficiency in addressing semantic heterophily in heterogeneous graph learning.

% \section*{Acknowledgments}
% This should be a simple paragraph before the References to thank those individuals and institutions who have supported your work on this article.


\bibliographystyle{IEEEtran}
\normalem
% \bibliography{reference}

% \newpage


\vfill

\end{document}